%% file: acl_latex.tex
\pdfoutput=1
\documentclass[11pt]{article}

\usepackage[final]{acl}
\usepackage{multirow}
\usepackage{times}
\usepackage{latexsym}
\usepackage{booktabs}
\usepackage{algorithm}
\usepackage{algpseudocode}
\usepackage{enumitem}

\usepackage[T1]{fontenc}

\usepackage[utf8]{inputenc}

\usepackage{microtype}
\usepackage{subcaption}
\usepackage[table,xcdraw]{xcolor}   
\usepackage{dsfont}
\usepackage{inconsolata}

\usepackage{graphicx}
\usepackage{kotex}
\usepackage[svgnames]{xcolor}
\usepackage{tcolorbox}
\usepackage{lipsum}
\usepackage{soul}
\input{math_commands}

\tcbset{
    colback=blue!5!white,
    colframe=blue!80!black,
    coltitle=white,
    fonttitle=\bfseries,
    boxrule=0pt,
    arc=4mm,
    outer arc=2mm,
}

\title{Safeguarding Privacy of Retrieval Data against Membership Inference Attacks: Is This Query Too Close to Home?}

\author{\quad \quad Yujin Choi \textsuperscript{\rm 1}\textsuperscript{\rm$\dagger$} \\ \hspace{-5em} \And \hspace{-3em}
  Youngjoo Park \textsuperscript{\rm 1}\textsuperscript{\rm$\dagger$} \hspace{-2em} \\  \And \hspace{-3em}
  Junyoung Byun \textsuperscript{\rm 2}\hspace{-1em} \\ \textsuperscript{\rm 1} Seoul National University, Republic of Korea \\
  \textsuperscript{\rm 2} Chung-Ang University, Republic of Korea 
  \\
  \textsuperscript{\rm 3} Korea Institute for Advanced Study, Republic of Korea \\ 
  \small\texttt{\{uznhigh, youngjoo0913, jaewook\}@snu.ac.kr, junyoungb@cau.ac.kr, jinseong@kias.re.kr}
  \hspace{-1em} \And
  \hspace{-3em} Jaewook Lee \textsuperscript{\rm 1}\hspace{-1em} \\ \And \hspace{-4em}
  Jinseong Park \textsuperscript{\rm 3}\thanks{Corresponding author. \textsuperscript{\rm $\dagger$}{Equal contribution.} }
  }

\begin{document}
\maketitle
\begin{abstract}

Retrieval-augmented generation (RAG) mitigates the hallucination problem in large language models (LLMs) and has proven effective for personalized usages. However, delivering private retrieved documents directly to LLMs introduces vulnerability to membership inference attacks (MIAs), which try to determine whether the target data point exists in the private external database or not. Based on the insight that MIA queries typically exhibit high similarity to only one target document, we introduce a novel similarity-based MIA detection framework designed for the RAG system. With the proposed method, we show that a simple detect-and-hide strategy can successfully obfuscate attackers, maintain data utility, and remain system-agnostic against MIA. We experimentally prove its detection and defense against various state-of-the-art MIA methods and its adaptability to existing RAG systems.
 \end{abstract}

\input{camera_ready}

\bibliography{custom}

\appendix
\input{manuscript_appendix}

\end{document}

%% file: math_commands.tex
\usepackage{amsmath,amsfonts,bm}

\def\eqref#1{equation~\ref{#1}}

\def\1{\bm{1}}

\def\0{{\bm{0}}}
\def\1{{\bm{1}}}

\DeclareMathAlphabet{\mathsfit}{\encodingdefault}{\sfdefault}{m}{sl}
\SetMathAlphabet{\mathsfit}{bold}{\encodingdefault}{\sfdefault}{bx}{n}



%% file: camera_ready.tex
\section{Introduction}
Large language models (LLMs) \cite{brown2020language, grattafiori2024llama} have demonstrated strength for general and common knowledge. However, they struggle to answer domain-specific or personalized questions, resulting in hallucinations that fabricate non-truth answers from the training set \cite{huang2025survey}. Retrieval-augmented generation (RAG) \cite{lewis2020retrieval} mitigates hallucination by giving information from external data retrieval into LLMs. RAG can provide reliable information by extracting relevant documents from an external database.

Membership inference attacks (MIAs) \cite{shokri2017membership} are malicious machine learning attacks that attempt to determine whether the target document was used in the training dataset.  Recent work has examined MIAs and extraction attacks on language models to study the privacy leakage \cite{shi2024detecting}. Even though MIAs achieve lower success rates on LLMs due to the massive size of training data samples \cite{puerto2024scaling}, privacy risks in RAG systems are significant \cite{zeng2024good}. As RAG retrieves a few top-$k$ documents and conveys them directly to LLMs, various attack methods \cite{liu2025mask,naseh2025riddle} have succeeded in inferring whether a target document is stored in a private retrieval database or not. 

However, safeguarding LLMs against MIA has no practical solution due to the huge model size or properties of language domains \cite{li2021membership}. Within RAG systems, the agent-based query filtering system is the only defense explored so far, but it is also broken by stealth attacks that evade detection \cite{naseh2025riddle}.
Thus, in this paper, we present a new safeguarding framework against MIAs by focusing on the observation that current MIAs on the RAG system rely on queries similar to only a single target document in the database.

In detail, we measure the similarity between an input query and the retrieved data points. Then, we check whether this similarity exceeds a threshold based on the Gumbel distribution \cite{gumbel1935valeurs} or not, which models the maximum values of data samples. For an input query that exceeds the threshold, indicating that the query is overly correlated with one specific document in the retrieval, we hide the data in the top-$k$ document conveyed to the LLMs.
We summarize our contributions as follows:
\begin{itemize}
    \item We propose a similarity-based method for detecting \underline{MI}A in \underline{RA}G systems using Gum\underline{bel} distribution, named \textbf{Mirabel}. To the best of our knowledge, this is the first attempt to study safeguarding strategies specifically against MIAs in RAG.
    \item Mirabel safeguards private external database against MIA through a simple detect-and-hide approach, which obfuscates attackers while preserving the utility of the RAG systems. 
    \item We empirically show the effectiveness of our method against various measures, and its comparability to existing defense methods.
\end{itemize}

\section{Related Works}
\subsection{Retrieval-Augmented Generation}
RAG is a strategy that enhances LLMs by integrating external data retrieval into the generation process \cite{lewis2020retrieval}. At its core, an RAG system comprises three primary components: an external database $\mathcal{D}$ of textual documents, a retriever $R$, and a generator (i.e., LLM) $G$. When a user submits a query $q$, the retriever identifies the top-$k$ contextual retrieval from $\mathcal{D}$ based on similarity, such as cosine similarity, often computed in an embedding model $\phi(\cdot)$ \cite{karpukhin2020dense}:
\begin{equation}
\label{eq:retrieval}
R_k(q)=\arg\mathrm{top\textit{k}} _{d\in\mathcal{D}}\mathtt{sim}(q,d).
\end{equation}
These retrievals are then combined with the original query to form an augmented context, which is passed to the generator. The generator produces an output based on the retrieval \cite{gao2023retrieval, shuster2021retrieval} as follows:
\begin{equation}
\label{eq:prompting}
p(q)=\mathtt{RAGprompt}(q, R_k(q)),
\end{equation}
\begin{equation}
\label{eq:generation}
\mathtt{response}=G(p(q)).
\end{equation}
RAG improves response accuracy and reduces hallucinations frequently observed in pure LLMs.
Furthermore, RAG offers architectural flexibility: any of the three core modules ($\mathcal{D}$, $R$, and $G$) can be replaced or updated independently without requiring end-to-end retraining \cite{cheng2023lift}. Moreover, query rewriting (e.g., correcting ambiguous) and specialized retrieval (e.g., token- or graph-based) can be used in RAG \cite{ram2023context}.

\begin{figure*}[t]
    \centering \vspace{-3mm}
    \begin{subfigure}[t]{0.44\linewidth}
        \centering
        \includegraphics[width=\linewidth]{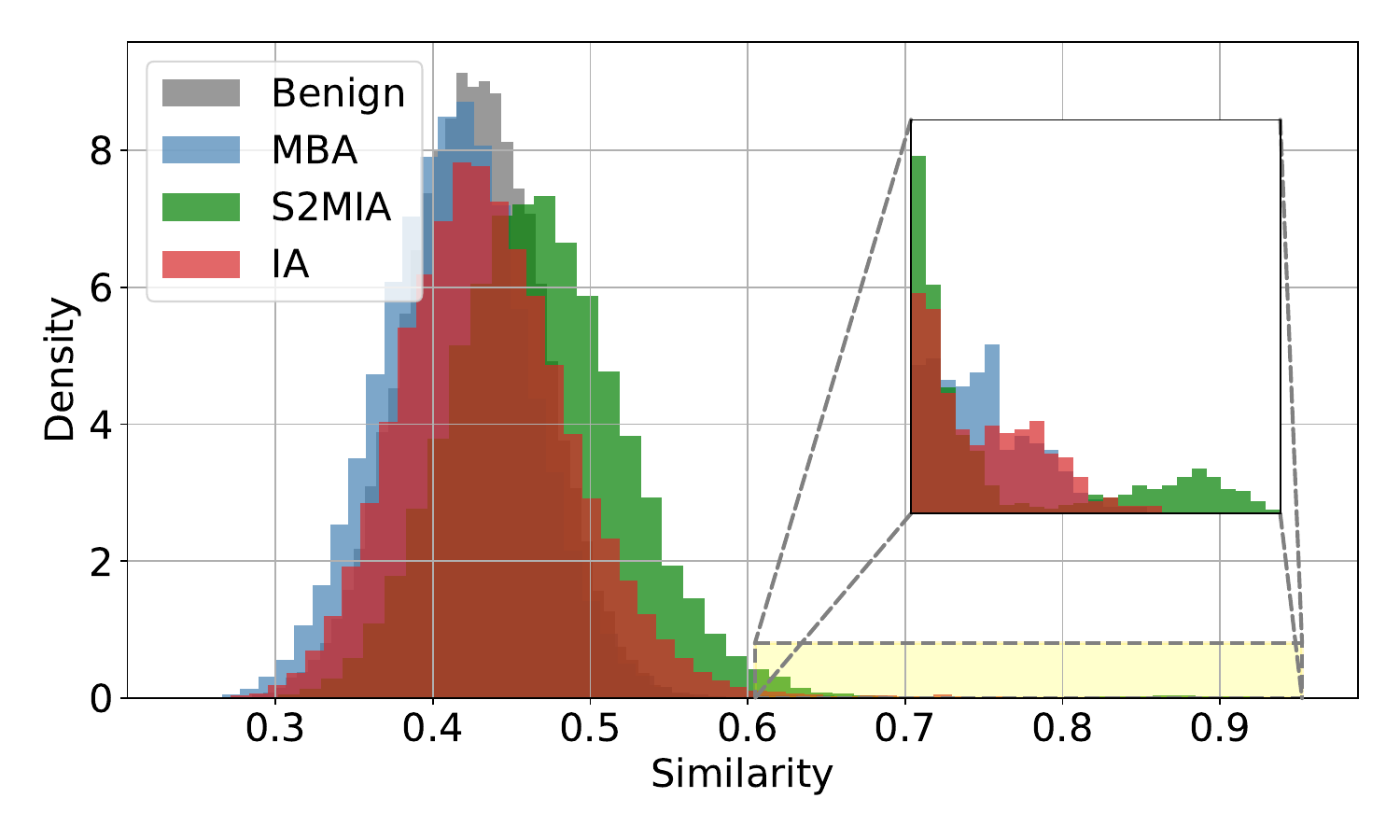}\vspace{-2mm}
        \caption{All similarities between query and candidates.}
        \label{fig:distribution_all}
    \end{subfigure}
    \begin{subfigure}[t]{0.44\linewidth}
        \centering
        \includegraphics[width=\linewidth]{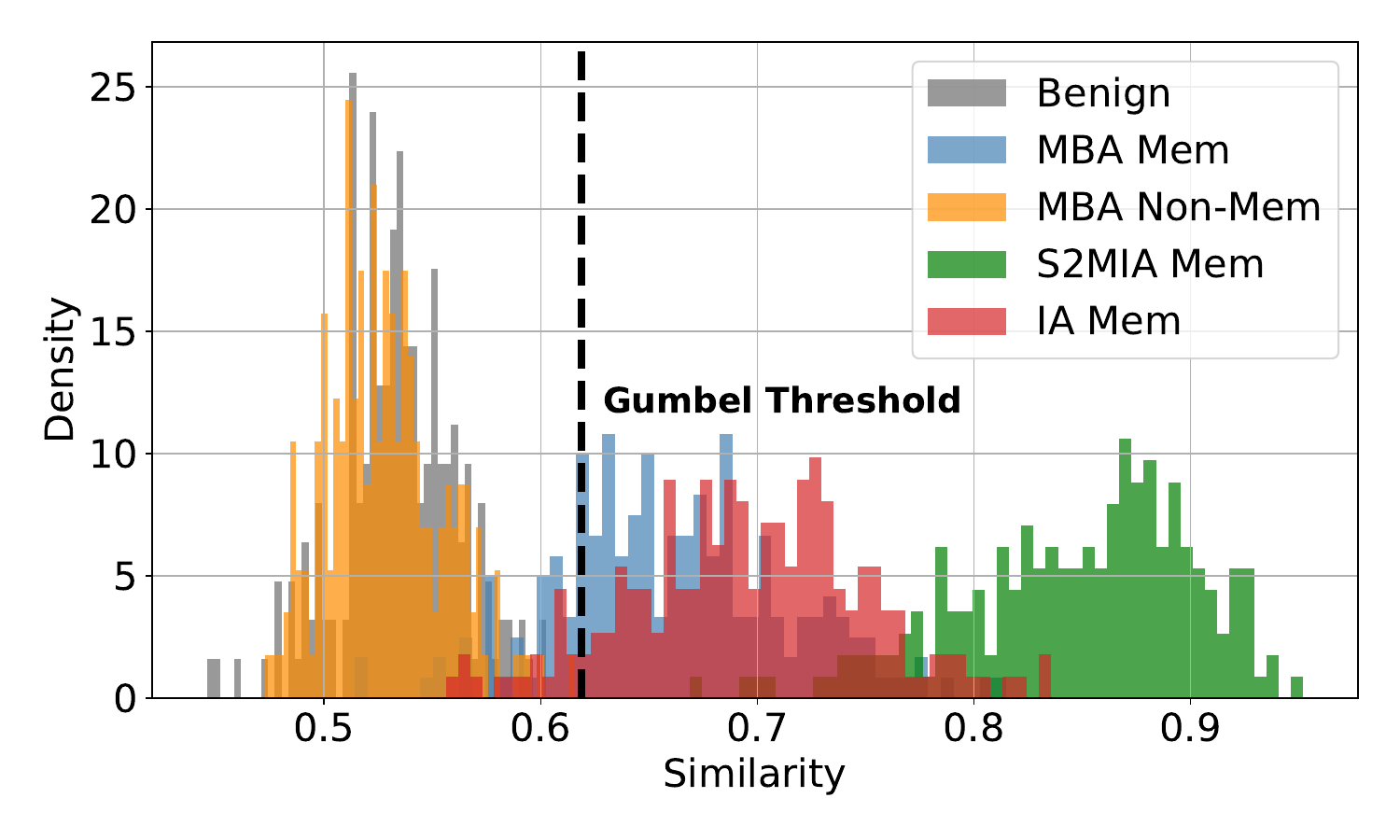}\vspace{-2mm}
        \caption{Top-1 similarity per query with Gumbel threshold.}
        \label{fig:distribution_top1}
    \end{subfigure}
    \caption{Distributions of similarity scores between queries and retrieved data. We visualize both the full similarity distributions and the top-1 similarities. A Gumbel-based threshold $\bigcup_q S_q$ is marked for reference.}\vspace{-3mm}
    \label{fig:distribution}
\end{figure*}

\subsection{Privacy Leakage of RAG System}
Despite the utility improvements, the RAG systems inevitably pose privacy concerns when dealing with sensitive or proprietary data in the retrieval. For example, when dealing with sensitive data, the use of RAG increases the risk of serious legal complications and leaks personal privacy if the documents in the external database for retrieval are exposed or attacked. Thus, we focus on the privacy of the private external database in this paper.

The purpose of MIAs for the RAG system is to determine whether a target document is stored in the database. Beginning with directly asking whether the document is in the database or not \cite{anderson2024my}, various attacks are proposed:
S$^{2}$MIA \cite{li2024generating} that provides the first half of the document and requests completion;
MBA \cite{liu2025mask} that prompts prediction of masked tokens; and
IA \cite{naseh2025riddle}, indicating an interrogation attack, that asks multiple queries that are hard to answer without the document.
All of the attacks then exploit the output of the RAG system to infer the membership of the target document.

\subsection{Safeguarding Attacks in RAG}
To safeguard against MIAs in RAG systems, \citet{naseh2025riddle} investigated a simple filtering method that asks an LLM agent, such as GPT-4o, to classify incoming queries as benign or malicious. However, this agent-based method struggles with several challenges that will be further discussed in Section \ref{sec:goals}.

As a complementary safeguard system, differential privacy (DP) provides a mathematically rigorous privacy guarantee for sensitive data. \citet{duan2023flocks} introduced a DP framework that hides private information through noisy labeling against MIAs. DP LLMs have also been explored for in-context learning \cite{tang2024privacypreserving} and private prompt tuning \cite{hong2024dpopt}. For the RAG system, DP-RAG
\cite{grislain2024rag} protects the privacy of data in the data retrieval, and DPVoteRAG \cite{koga2024privacy} uses private voting. 
\citet{tran2025tokens} used a machine unlearning to delete highly memorized tokens, but their approach targets fine-tuning LLMs and is not focused on RAG privacy.

\section{Scenarios}
\label{sec:scenarios}

Following the previous studies \cite{liu2025mask,naseh2025riddle}, we assume three parties in our scenario: 1) the \textbf{operator} of the RAG system with \textbf{private external database}, 2) the  \textbf{benign users} of the RAG, and 3) the malicious \textbf{attackers} attempting MIAs to the RAG system. For instance,

\noindent
\emph{
``A healthcare AI operator deploys a medical-diagnosis chatbot that retrieves private patient records. Benign users consult the chatbot to assess their health, but attackers attempt MIAs. The operator must preserve privacy against attackers while providing accurate answers to benign users." }
\paragraph{Types of Queries}
Based on the scenario, we divide the queries for RAG systems into three types:
member attack query $q^m_a$ by attackers of malicious questions in the external database, non-member attack query $q^n_a$ by attackers of malicious questions but not in the external database, and benign query $q_b$ of benign users without malicious intention.
\paragraph{Details about Attacker}
Given a target document $d$, an attacker aims to decide whether $d$ is contained in the external database $\mathcal{D}$ by classifying its associated query $q_a$ as a member attack query $q^m_a$ or a non-member attack query $q^n_a$. We assume the attacker cannot access $\mathcal{D}$ or the parameters of LLMs.

\paragraph{Details about Operator's Defenses}
The defense should prevent attackers from recognizing the membership information while delivering correct responses to benign users without prior knowledge of attacks, such as their prompts or patterns. The defender also cannot access LLM parameters, just relying on access to the external database $\mathcal{D}$ and its embedding model $\phi$. Each query must be answered immediately without pending the next queries.


\subsection{Motivation and Goals}
\label{sec:goals}
To safeguard against MIAs, it is essential to determine whether an input query is malicious to the RAG system for the private external database. Recently, \citet{naseh2025riddle} proposed an LLM agent-based detection method to evaluate the malicious intent of input queries, independent of the external database. To judge either benign queries $q_b$ or attack queries $q_a$, the agent identifies harmful or extraction phrases within queries. However, \citet{naseh2025riddle} also showed that the agent-based detection can be deceived by stealth queries, which are crafted attack queries designed to mimic benign queries. Moreover, simply rejecting all suspected attack queries may inadvertently reveal to attackers filtering phrases and detour the detection systems.

These limitations suggest that an effective defense requires not simply blocking queries but obfuscating the attacker's knowledge. An attack succeeds only if it can correctly separate $q_a^m$ from $q_a^n$ using the RAG's responses. Hence, to defend against the attack, the responses should be made statistically indistinguishable. Achieving this in turn demands that the defender distinguishes between $q_a^m$ and $q_a^n$, so the responses can be selectively perturbed. Therefore, we set two complementary goals: 1) accurately separate member attack queries from non-member attack queries, and 2) respond with the corresponding responses so that an attacker cannot classify the member and non-member cases.

\section{Proposed Method} \label{sec:detection}
\subsection{Is This Query Too Close to Home? \qquad MIA Detection with Gumbel Distribution}
By analyzing the similarity of queries, we propose a detection method for the RAG system that distinguishes member attack queries ($q^m_a$) from other types of queries ($q_b$ and $q^n_a$). 
Specifically, our goal is to find a threshold $\tau$ to classify a query as a member attack $q_a^m$ if the maximum similarity exceeds:
\begin{equation}
\max_{d \in \mathcal{D}} \mathtt{sim}(q,d) > \tau.
\end{equation}
\noindent
To find the threshold $\tau$, we focus on the distributional differences between member attack queries and other queries \cite{wen2024membership}.

Figure \ref{fig:distribution_all} demonstrates $\bigcup_q S_q$, the cosine similarities between all queries and the external database, where $S_q = \{\mathtt{sim}(q,d)|d\in \mathcal{D}\}$ is a set of similarities between a query $q$ and the total database. 
The figure shows (i) similarities of the benign query $q_b$ and the total database follow a (approximately) normal distribution, and (ii) for the member attack queries $q_a^m$, the distribution appears similar to $q_b$ in the most likely region, but exhibits extreme values in the right tail, which are too similar with a private external database.

Based on these observations, we propose a detection method using the Gumbel distribution, which represents the maximum value of data samples \cite{gumbel1935valeurs}. In previous works on LLM privacy, the Gumbel distribution is widely used to find top-$k$ selection \cite{durfee2019practical,hong2024dpopt}.
By extreme value theory, if $n$ random variables are sufficiently large and follow an i.i.d. normal distribution, i.e.,
\begin{align}
X_1, \cdots, X_n \overset{\text{i.i.d.}}{\sim} \mathcal{N}(\mu_q, \sigma_q^2), 
\end{align}
for some $\mu_q$ and $\sigma_q$, then the maximum of the samples converges in distribution to a Gumbel distribution:
\begin{align}
\max\{X_1, \cdots, X_n\} \xrightarrow{d} {Gumbel}(\mu_n, \beta_n),
\end{align}
where $\beta_n = \sigma/\sqrt{2\ln n}$ and 
\begin{align}\label{eq:gumbel_mean}
    \mu_n &= \mu_q + \sigma_qd \sqrt{2\ln n - \ln(\ln n) - ln(4\pi)} \nonumber\\
    &\approx \mu_q + \sigma_q d \sqrt{2\ln n} \quad (\text{as }n\to\infty).
\end{align}


\noindent
Under the observations (i) and (ii) in Figure \ref{fig:distribution_all}, we assume that the samples from $S_q\!\setminus\!\{s_{\max}\}$ follow a normal distribution for each $q$, where $s_{\max}$ is the maximum similarity in $S_q$. 
Let $\mu_q$ and $\sigma_q$ be the mean and standard deviation of $S_q\!\setminus\!\{s_{\max}\}$. Then, the threshold $\tau$ can be calculated as follows:
\begin{align}\label{eq:threshold}
    \tau = \mu_n + c \cdot \sigma_q / \sqrt{2\ln n},
\end{align}
where 
$c = -\ln(-\ln(1-\rho))$ is a critical value of the Gumbel distribution for significance level $\rho$.

Figure~\ref{fig:distribution_top1} shows histograms of the maximum value of each $S_q$, i.e., $s_{\max}$ for each query.
The dashed line represents the threshold based on a Gumbel distribution, which was computed from
$\bigcup_{q_b} S_{q_b}$.  
Thresholds computed for each attack type for separate $s_{\max}$ for non-member attack query $q_a^n$ and member-attack query $q_a^m$ are provided in Figure \ref{fig:toy_nonmember} in Appendix \ref{app:add_exp}. 
Most $s_{\max}$ values of the member attack queries exceed the Gumbel-based threshold, whereas those of benign and non-member attack queries remain below it. These findings suggest that $s_{\max}$ of member attack queries $q_a^m$ can be successfully separated from $q_b$ and $q_a^n$ by the Gumbel-based threshold.


To this end, we name the proposed detection method as \textbf{Mirabel}, i.e., \underline{MI}A in \underline{RA}G systems using Gum\underline{bel}. Mirabel not only detects whether the query $q$ is a member attack, but also identifies the specific document that is likely the target. Moreover, it can be incorporated into the standard RAG system naturally, as it uses the similarity scores already computed for top-$k$ selection. 
The detailed method is shown in Algorithm \ref{alg:detection}.

\begin{table}[!t]
\centering
\resizebox{0.95\linewidth}{!}{%
\begin{tabular}{l|c|cccc}
\toprule
\textbf{Test set} & \textbf{Benign} & \textbf{MBA} & \textbf{S$^{2}$MIA} & \textbf{IA} \\
\midrule
$S_q$                               & 0.469  & \underline{0.027}$^*$ & \underline{0.001}$^*$ & \underline{0.012}$^*$ \\
$S_q\!\setminus\!\{s_{\max}\}$          & 0.400 & 0.511 & 0.293        & 0.226 \\
\bottomrule
\end{tabular}}
\caption{Average $p$-values of the normality test on the total data set ($S_q$) and on the set with the maximum similarity removed ($S_q\!\setminus\!\{s_{\max}\}$). * : $p$-value < 0.05.}
\label{tab:avg-pvalues}
\end{table}

\begin{algorithm}[!t]
\caption{Proposed Mirabel Detection}
\label{alg:detection}
\begin{algorithmic}[1]
\Statex \textbf{Input:} Query embedding function $\phi$, corpus embeddings $\{e_d = \phi(d) : d \in \mathcal{D}\}$, query $q$, significance level $\rho$ 
\Statex \textbf{Output:} detection result, target document $d_t$
\Statex \textbf{Initialization: }  $S_q \leftarrow \{\}$, $e_q \leftarrow \phi(q)$
\ForAll{$d \in \mathcal{D}$}
    \State $s_d \leftarrow \cos(e_q, e_d)$ \Comment{Compute \texttt{sim} ($q, d$)}
    \State $S_q \leftarrow S_q \cup \{s_d\}$
\EndFor
\State $s_{\text{max}}, d_t \leftarrow \max(S_q)$  \Comment{Find (arg)max \texttt{sim}}
\State $\mu_q, \sigma_q \leftarrow \text{mean}(S_q\!\setminus\!\{s_{\max}\}), \text{std}(S_q\!\setminus\!\{s_{\max}\})$
\State $\mu_n \leftarrow \mu_q + \sigma_q \cdot \sqrt{2\ln n}$ \Comment{Gumbel mean}
\State $c \leftarrow - \ln(-\ln (1-\rho))$ \Comment{Critical value}
\State $\tau \leftarrow \mu_n + c \cdot \sigma_q / \sqrt{2\ln n}$ \Comment{Threshold}
\State \textbf{return} ($s_{\max} > \tau$), $\mathds{1}[s_{\max} > \tau] \cdot d_t$
\end{algorithmic}
\end{algorithm}\vspace{-2mm}

\label{sec:method}
\begin{figure*}[!t]
    \centering \vspace{-2mm}
    \includegraphics[width=\textwidth] {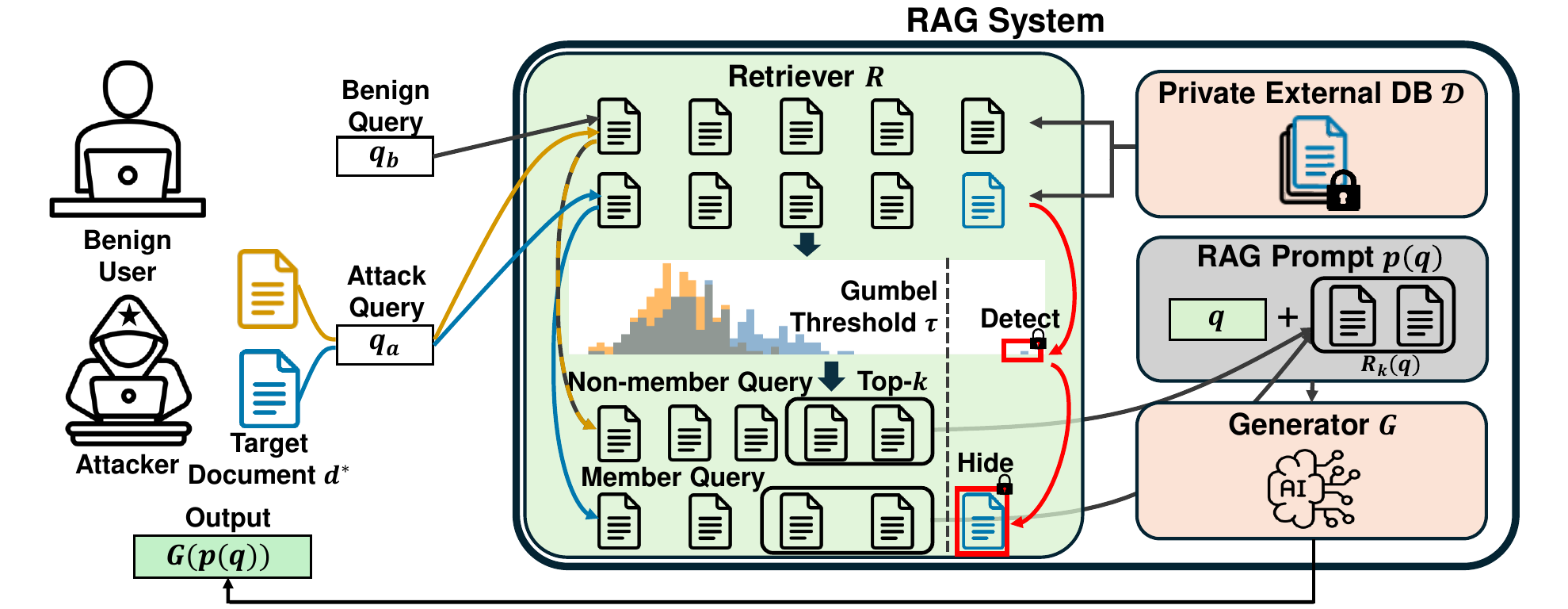}\vspace{-1mm}
    \caption{Illustration of our proposed Mirabel. We perform our detection to classify whether an input query is a member attack query $q_a^m$. If it detected as $q_a^m$, we hide it from data retrieval and proceed standard RAG system.} 
    \label{fig:proposed_method} \vspace{-1mm}
\end{figure*}
\paragraph{Normality test} 
To validate our assumption, we conduct a normality test on $S_q$ and $S_q\!\setminus\!\{s_{\max}\}$.
Table \ref{tab:avg-pvalues} demonstrates the mean $p$-values from D'Agostino and Pearson’s normality test, averaged across all queries $q$. 
Across all MIAs, the $p$-values of total similarity scores in $S_q$ are less than 0.05, indicating that we can reject the hypothesis that the scores follow a normal distribution. 
In contrast, for a benign query $q_b$, $p > 0.05$, we cannot reject that $S_{q_b}$ follows a normal distribution.
 

In contrast, when the maximum similarity score $s_{\max}$ is removed from $S_q$, we observe $p>0.05$ for all queries, indicating that we cannot reject that $S_q\!\setminus\!\{s_{\max}\}$ follows a normal distribution.
As removing $s_{\max}$ indicates the target document is removed in the database, we cannot reject the hypothesis of normality even for non-member attack queries $q_a^n$.

The distributional difference between $S_q$ and $S_q \setminus \{s_{\max}\}$ indicates that $s_{\max}$ is an extreme outlier relative to values drawn from a normal distribution for $q_a^m$.
Since $s_{\max}$ is distinctive in the Gumbel distribution, this supports the proposed Mirabel method, which identifies MIAs for the RAG system by leveraging the confidence interval of the Gumbel distribution.

\subsection{Detect-and-hide to Defend MIA} 

Leveraging the proposed Mirabel for MIA detection, we propose a simple defense method that safeguards the data sample in the external database with a simple \textit{detect-and-hide} strategy. In the detection, the query is evaluated using Mirabel to determine whether it is a member attack query $q_a^m$. In the hiding, if the query $q$ is classified as $q_a^m$, the target document identified by Mirabel is hidden from the external database, and the standard RAG system proceeds with the remaining documents. If the query is not classified as a member attack, the standard RAG system proceeds without any modification.
Figure~\ref{fig:proposed_method} illustrates our proposed defense method.


First of all, this simple defense confuses the attacker about whether the RAG output is from members or non-members, since the response of the retriever no longer includes the target document in both cases. Upon the observation of the normality in the previous subsection, the response distribution remains indistinguishable between member and non-member queries.

Our defense method, furthermore,  preserves utility for benign queries $q_b$. When the query is not detected as $q_a^m$, the system behaves exactly the same as the original RAG system. Unlike other strong privacy-preserving mechanisms (e.g., differential privacy), which may degrade utility even for benign queries, our approach introduces minimal utility loss when the query is not detected as $q_a^m$.

Finally, our safeguarding strategy is agnostic to any RAG systems, as the detection only requires similarity scores, which is a basic part of the RAG system to calculate the top-$k$ documents in the external database.
Thus, it is easy to apply and does not require any modification of the system. The only additional time is to compute the Gumbel distribution and hide the corresponding document from the database for retrieval if the query is classified as $q_a^m$. 
Based on this advantage, we show that integrating our method into existing privacy-preserving systems results in better defenses.

\section{Experiments}
\subsection{Experimental Setups}

\paragraph{Dataset}
To evaluate detection and defense performance against MIAs, we utilize three datasets: NFCorpus, SciDOCS, and TREC-COVID, drawn from the BEIR benchmark \cite{thakur2021beir}, as employed in \cite{naseh2025riddle}. These corpora represent sensitive scientific and medical domains.
In addition to evaluating the utility of the RAG system on benign queries, we followed \cite{koga2024privacy} and used two question-answer benchmark datasets: Natural Questions (NQ) \cite{kwiatkowski2019natural} and TriviaQA \cite{joshi2017triviaqa}. For both datasets, Wikipedia serves as the external database \cite{lewis2020retrieval}.

\paragraph{Baselines}
We evaluate performance for our detection and defense method using three MIA methods: S$^2$MIA \cite{li2024generating}, MBA \cite{liu2025mask}, and IA \cite{naseh2025riddle}. S$^{2}$MIA feeds the first half of the target document to the RAG and scores membership with BLEU and perplexity against the full text. MBA masks tokens and counts how many the generator recovers. IA lets an LLM craft 30 inference questions per document, and correctness on them is the score. 

For the comparison method of detection, we perform an agent-based detection using GPT-4o to distinguish between following the approach of \cite{naseh2025riddle}, detailed in Appendix \ref{app:prompt}.

As our work is the first work for defending MIAs for RAGs, we choose to compare with the most well-known privacy notion, DP \cite{dwork2006calibrating}, which guarantees indistinguishability of individual data in the private external database. 
For DP-RAG \citep{grislain2024rag}, we inject the noise in the next token prediction based on DP in-context learning \cite{tang2024privacypreserving} without considering the privacy of top-$k$ selection \cite{koga2024privacy}. 

For the privacy budget $\epsilon$, we set $\epsilon = 2$ for base DP-RAG. To compare with a relaxed privacy setting, we also set $\epsilon=100$ and referred to it as DP-RAG Large (in short, DP-RAG-L).

\paragraph{Metrics}
For the detection component, we measure accuracy, precision, recall, and F1-score to determine how effectively the system identifies queries targeting member documents. 

To assess the utility of the system, we measure (i) the exact match (EM), which indicates the proportion of generated responses containing the gold labels with top-$k$ retrievals, and (ii) R@k, which indicates the proportion of retrievals containing the gold labels among the top-$k$ results.

For defense, we suggest the measures of \textbf{attack resistance} and \textbf{indistinguishability}. The attack resistance is measured based on the accuracy of the attack. However, in MIA, since it is a binary classification, accuracy alone may underestimate attack performance, especially when it falls below 0.5. We instead define adjusted attack accuracy as:
\begin{equation}
\label{eq:attack}
    \textit{Adjusted attack accuracy} = \text{max(acc, 1-acc)} - 0.5.
\end{equation}
This metric reflects the adversary's advantage over random guessing (0.5 in a balanced setting) and ranges from 0 to 0.5.  Lower values indicate weaker attack success and, consequently, stronger defense.

To evaluate indistinguishability, we use the Kolmogorov–Smirnov (KS) test \cite{an1933sulla} to compare the distributions of responses for member and non-member queries. The KS statistic measures the maximum difference between the two empirical distributions; a smaller value indicates greater similarity, and thus higher indistinguishability.


\paragraph{RAG Setting and Implementation}
We employed a 1024-dimensional BGEm3 embedder \cite{li2023towards} to find the similarity of queries and documents in the retriever. In our defensible retriever, we use a Mirabel based on the significance parameter $\rho$, to decide whether to filter documents before extracting top-$k$ candidates via cosine similarity. In our experiment, we set the significance level $\rho = 0.05$. In evaluations concerning MIA detection and defense on NFCorpus, SCIDOCS, and TREC-COVID, we set $k=3$. For utility evaluations on the Wikipedia corpus, we tested $k\in\{5,20,100\}$. 

For the generator LLMs, we adopted Llama 3.2 3B Instruct and Llama 3.1 8B Instruct \cite{grattafiori2024llama}. Specifically, we used Llama 3.1 for the utility evaluations and Llama 3.2 for the remaining evaluations. Further details on the RAG prompt can be found in the Figure \ref{ex:rag_prompt}. Experiments were mainly conducted on a single A100 or H100 GPU with 96 GB VRAM.

\noindent
Additional details are shown in Appendix \ref{app:exp}.

\begin{table*}[!t]
\centering
\resizebox{0.95\textwidth}{!}{%
\begin{tabular}{lc|ccc>{\columncolor[HTML]{E5FFE5}}c|ccc>{\columncolor[HTML]{E5FFE5}}c}
\toprule
&                       & \multicolumn{4}{c}{Agent-based Detection} & \multicolumn{4}{c}{Mirabel Detection} \\ 
\cmidrule(lr){3-6}\cmidrule(lr){7-10}
{Attacks} & {Data} & Acc ($\uparrow$) & F1 ($\uparrow$) & Precision ($\uparrow$) & Recall ($\uparrow$) & Acc ($\uparrow$) & F1 ($\uparrow$) & Precision ($\uparrow$) & Recall ($\uparrow$) \\ 
\midrule
\multirow{3}{*}{S$^{2}$MIA} & NF   & 0.715 & 0.772 & 0.644 & 0.962 & 0.844 & 0.865 & 0.762 & \textbf{1.000} \\
                         & SCI  & 0.726 & 0.780 & 0.651 & 0.974 & 0.853 & 0.871 & 0.776 & \textbf{0.992} \\
                         & TREC & 0.745 & 0.792 & 0.669 & \textbf{0.970} & 0.880 & 0.886 & 0.846 & 0.930 \\ \cmidrule{1-10}
\multirow{3}{*}{MBA}     & NF   & 0.727 & 0.783 & 0.650 & 0.986 & 0.847 & 0.867 & 0.766 & \textbf{1.000 }\\
                         & SCI  & 0.730 & 0.785 & 0.652 & 0.986 & 0.855 & 0.873 & 0.775 & \textbf{1.000 }\\
                         & TREC & 0.745 & 0.794 & 0.667 & 0.980 & 0.995 & 0.995 & 1.000 & \textbf{0.990} \\ \cmidrule{1-10}
\multirow{3}{*}{IA}      & NF   & 0.482 & 0.012 & 0.125 & 0.006 & 0.799	& 0.750	& 0.896 &	\textbf{0.817}\\
                         & SCI  & 0.485 & 0.030 & 0.258 & 0.016 &0.827	&0.772	&0.928	&\textbf{0.843}\\
                         & TREC & 0.500 & 0.000 & 0.000 & 0.000 & 0.720&	0.744	&0.670	&\textbf{0.705} \\
\bottomrule
\end{tabular}}
\caption{Detection performance of Mirabel compared to agent-based detection. For simplicity, we denote NFCorpus as NF, SCIDOCS as SCI, and TREC-COVID as TREC throughout the following tables.} \vspace{-2mm}
\label{tab:detection}
\end{table*}

\subsection{Detection Evaluation}

Table \ref{tab:detection} presents the member attack query detection performance compared to agent-based methods. To detect member attack queries, $q_a^m$ was labeled as 1 and otherwise 0. Note that the agent-based detection \cite{naseh2025riddle} labels both attack queries $q_a^m$ and $q_a^n$ as 1, which cannot distinguish attack queries only. 
To balance the binary classification, we set the total number of $q_b$ and $q_a^n$ to be the same as $q_a^m$.

Mirabel detection demonstrates stable performance across all MIAs, including the IA with stealth queries, which the agent-based detection struggles to detect. 
Notably, our method achieved a high recall score, indicating successful detection of $q_a^m$, while its precision was slightly lower due to misclassifying $q_b$ as $q_a^m$. 
In other words, our method exhibits a lower Type II error, which is generally considered more critical in attack detection. 
However, despite this minor reduction, our precision remains higher than that of agent-based detection.



The agent-based methods cannot classify a query as related to private members or not.
In the next section, we will analyze how these errors impact both attack performance and utility. 
Additional metrics, including detection performance comparing only $q_b$ and $q_a^m$, are provided in Appendix \ref{app:add_exp}. The results show that our method achieves a high recall score. 




\subsection{Defense Evaluation}
Safeguarding against MIA has two main goals: (i) reducing attack performance and (ii) preserving the utility of RAG systems, while preventing the attacker from gaining private information. Thus, we must measure both attack degradation and utility preservation for defense methods. 

\paragraph{Utility Preservation}

\begin{table}[]
\resizebox{0.99\linewidth}{!}{%
\begin{tabular}{ll|ccc|ccc}\toprule
\multicolumn{1}{c}{} & \multicolumn{1}{c}{} & \multicolumn{3}{c}{EM ($\uparrow$)} & \multicolumn{3}{c}{R ($\uparrow$)} \\ 
\cmidrule(lr){3-5}\cmidrule(lr){6-8}
Data & \multicolumn{1}{c}{System} & \multicolumn{1}{c}{@5} & @20 & \multicolumn{1}{c}{@100} & @5 & @20 & @100 \\ \midrule
NQ   & RAG     & 0.272 & 0.263 & 0.313 & 0.222 & 0.354 & 0.474 \\
     & DP-RAG  & 0.030 & 0.051 & 0.020 &   –   &   –   &   –   \\
     & Ours    & 0.253 & 0.253 & 0.273 & 0.172 & 0.303 & 0.443 \\ \midrule
TRIV & RAG     & 0.730 & 0.725 & 0.755 & 0.575 & 0.705 & 0.840 \\
     & DP-RAG  & 0.255 & 0.225 & 0.240 &   –   &   –   &   –   \\
     & Ours    & 0.700 & 0.725 & 0.740 & 0.515 & 0.675 & 0.825 \\ 
\bottomrule
\end{tabular}}
\caption{Utility measures computed with benign queries.}
\label{tab:utility}
\vspace{-2mm}
\end{table}


We first demonstrate that our method largely avoids the utility degradation associated with the existing privacy-preserving approach for RAG systems, DP-RAG, in Table \ref{tab:utility}.


EM@k, which measures the quality of the generated answers, shows that our method performs comparably to the standard RAG, while DP-RAG results in greater utility degradation.

In contrast, R@k, which measures how well the system retrieves relevant documents, shows a moderate decrease compared to the original RAG system. This is because our detection method has a slightly larger Type I detection error.

Despite degradation in R@k, our method maintains high performance in EM, because it retains the remaining top-$k$ documents (i.e., from top-2 to top-($k+1$)) by hiding only the target member document. Since benign queries do not rely on a single document but instead reference multiple relevant documents, we can mitigate the utility loss. 


\paragraph{Attack Resistance}

To evaluate the performance of our defense, we measure the attack performances of MIAs in Table \ref{tab:adv_acc}. 
Without defense, MIAs achieved high adjusted attack accuracy, indicating the attacks are successful. However, after applying the defenses, the accuracy decreases, suggesting that the attacker struggles to distinguish whether the response is from a member or a non-member.
Our method effectively reduces the attack accuracy and achieves performance comparable to DP-RAG (-L), which is considered a strong defense despite its inherent privacy-utility trade-off.

From the perspective of detection failure, attacker performance is related to the Type II error. Type II errors occur when the detection method fails to identify a member query, preventing the removal of the target document. This can increase the number of true positives, as the MIA will classify members correctly as members.

Since our detection method has a low Type II error rate, it results in a strong defense. However, in the case of the IA on the TREC-COVID dataset, even with a low recall and high Type II error, the defense performance remains strong.

This is because the Type II error occurs when the similarity score between the target document $d$ and the attack query $q_a^m$ is not significantly high.
In IA, such low similarity indicates that the attack query was either a general question related to multiple documents or not strongly associated with the target document. In these cases, the attack itself is also likely to fail.

\begin{table}[!t]
\centering
\small
\begin{tabular}{ll|rrr>{\columncolor[HTML]{E5FFE5}}r}\toprule
 &  & \multicolumn{1}{c}{NF} & \multicolumn{1}{c}{SCI} & \multicolumn{1}{c}{TREC} & \multicolumn{1}{>{\columncolor[HTML]{E5FFE5}}c}{Avg} \\ \midrule
\multicolumn{1}{c}{S$^2$} & RAG       & 0.188 & 0.126 & 0.119 & 0.144 \\
                          MIA & DP-RAG  & 0.021 & 0.015 & 0.004 & \textbf{0.013} \\
                          & DP-RAG-L  & 0.034 & 0.000 & 0.004 & \textbf{0.013} \\
                          & Ours      & 0.024 & 0.006 & 0.021 & 0.017 \\ \midrule
\multicolumn{1}{c}{MBA}   & RAG       & 0.377 & 0.406 & 0.313 & 0.365 \\
                          & DP-RAG  & 0.004 & 0.040 & 0.003 & \textbf{0.016} \\
                          & DP-RAG-L  & 0.014 & 0.036 & 0.184 & 0.078 \\
                          & Ours      & 0.019 & 0.101 & 0.139 & 0.086 \\ \midrule
\multicolumn{1}{c}{IA}    & RAG       & 0.403 & 0.380 & 0.254 & 0.346 \\
                          & DP-RAG  & 0.038 & 0.288 & 0.174 & 0.167 \\
                          & DP-RAG-L  & 0.267 & 0.305 & 0.020 & 0.197 \\
                          & Ours      & 0.008 & 0.010 & 0.083 & \textbf{0.034} \\ \bottomrule
\end{tabular}
\caption{Adjusted attack accuracy ($\downarrow$). A smaller value indicates weaker attack success, thus stronger defense.}
\label{tab:adv_acc}
\end{table}

\paragraph{Indistinguishability}

\begin{table}[!t]
\centering
\small
\begin{tabular}{ll|rrr>{\columncolor[HTML]{E5FFE5}}r}\toprule
 &  & \multicolumn{1}{c}{NF} & \multicolumn{1}{c}{SCI} & \multicolumn{1}{c}{TREC} & \multicolumn{1}{>{\columncolor[HTML]{E5FFE5}}c}{Avg} \\ \midrule
\multicolumn{1}{c}{S$^2$} & RAG       & 0.358 & 0.195 & 0.203 & 0.252 \\
                         MIA & DP-RAG  & 0.077 & 0.051 & 0.038 & 0.055 \\
                          & DP-RAG-L  & 0.060 & 0.035 & 0.058 & 0.051 \\
                          & Ours      & 0.057 & 0.039 & 0.043 & \textbf{0.046} \\ \midrule
\multicolumn{1}{c}{MBA}   & RAG       & 0.754 & 0.813 & 0.625 & 0.731 \\
                          & DP-RAG  & 0.022 & 0.081 & 0.015 & \textbf{0.039} \\
                          & DP-RAG-L  & 0.036 & 0.073 & 0.026 & 0.045 \\
                          & Ours      & 0.038 & 0.202 & 0.283 & 0.174 \\ \midrule
\multicolumn{1}{c}{IA}    & RAG       & 0.805 & 0.771 & 0.555 & 0.710 \\
                          & DP-RAG  & 0.088 & 0.418 & 0.091 & 0.199 \\
                          & DP-RAG-L  & 0.301 & 0.414 & 0.223 & 0.313 \\
                          & Ours      & 0.100 & 0.081 & 0.172 & \textbf{0.118} \\ \bottomrule
\end{tabular}
\caption{KS statistics ($\downarrow$) for S$^2$MIA, MBA, and IA scores. For S$^2$MIA not having a direct score, we report the average KS statistics of similarity and perplexity.}\vspace{-1mm} 
\label{tab:ks_all}
\end{table}

\begin{figure*}[htbp]
    \centering
    \begin{subfigure}[b]{0.33\textwidth}
        \centering
        \includegraphics[width=\linewidth]{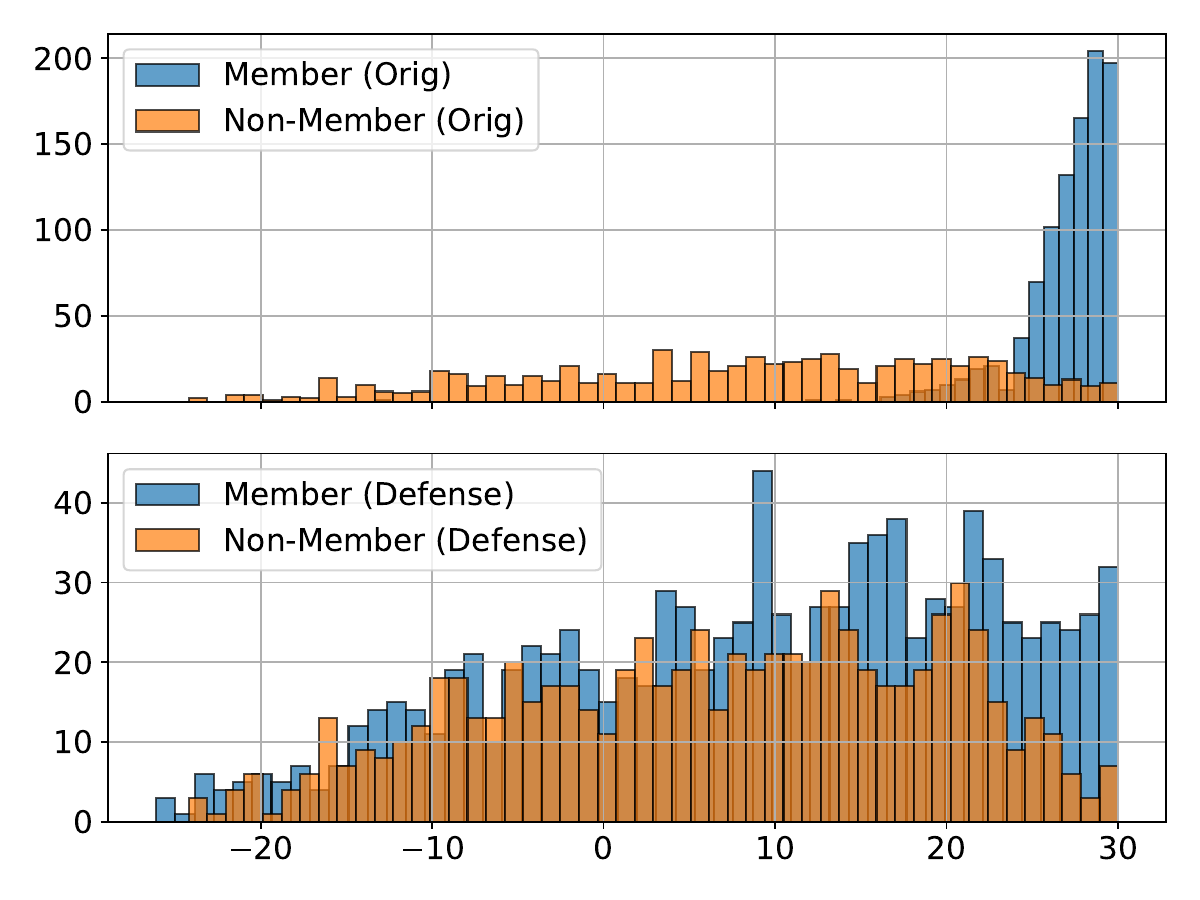}
        \caption{NFCorpus dataset.}
        \label{fig:nfcorpus_indisting}
    \end{subfigure}
    \begin{subfigure}[b]{0.33\textwidth}
        \centering
        \includegraphics[width=\linewidth]{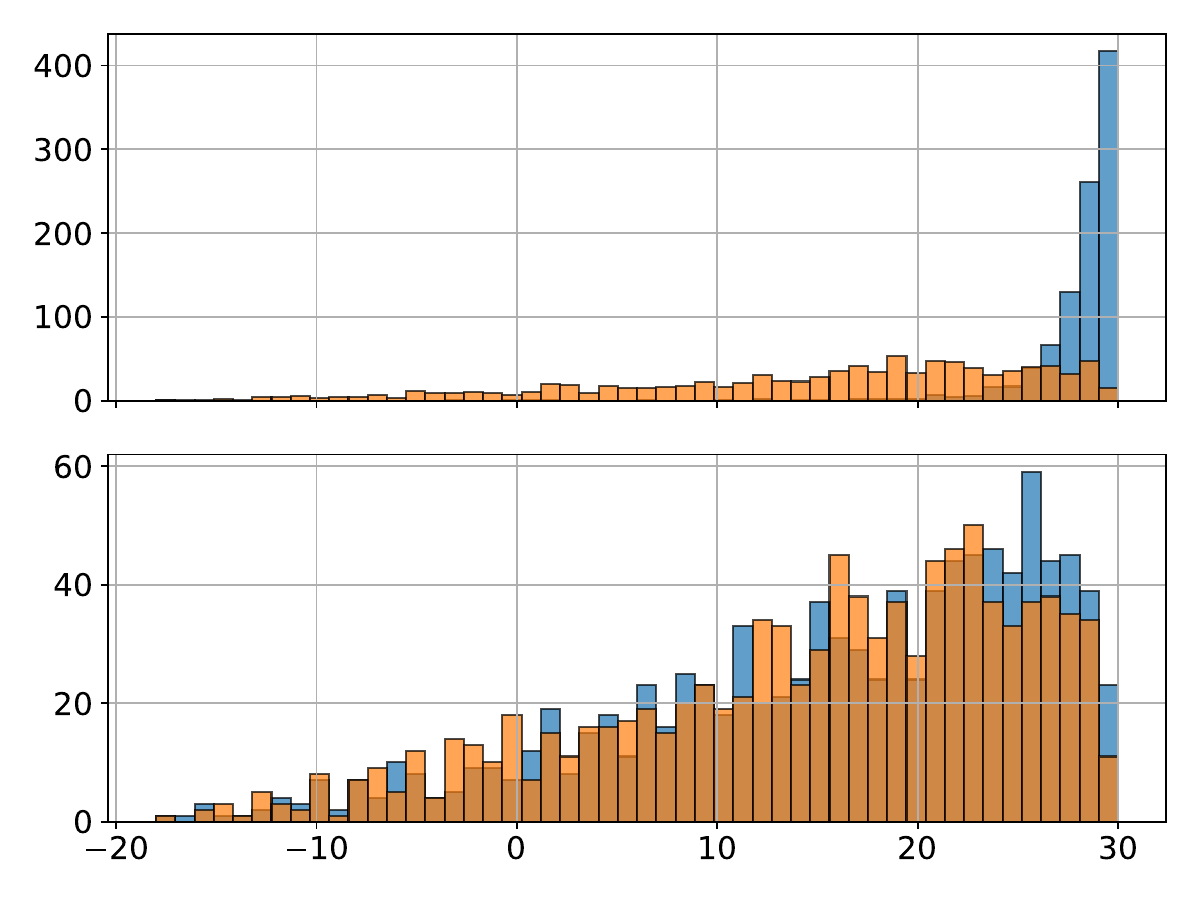}
        \caption{SCIDOCS dataset.}
        \label{fig:scidocs_indisting}
    \end{subfigure}%
    \begin{subfigure}[b]{0.33\textwidth}
        \centering
        \includegraphics[width=\linewidth]{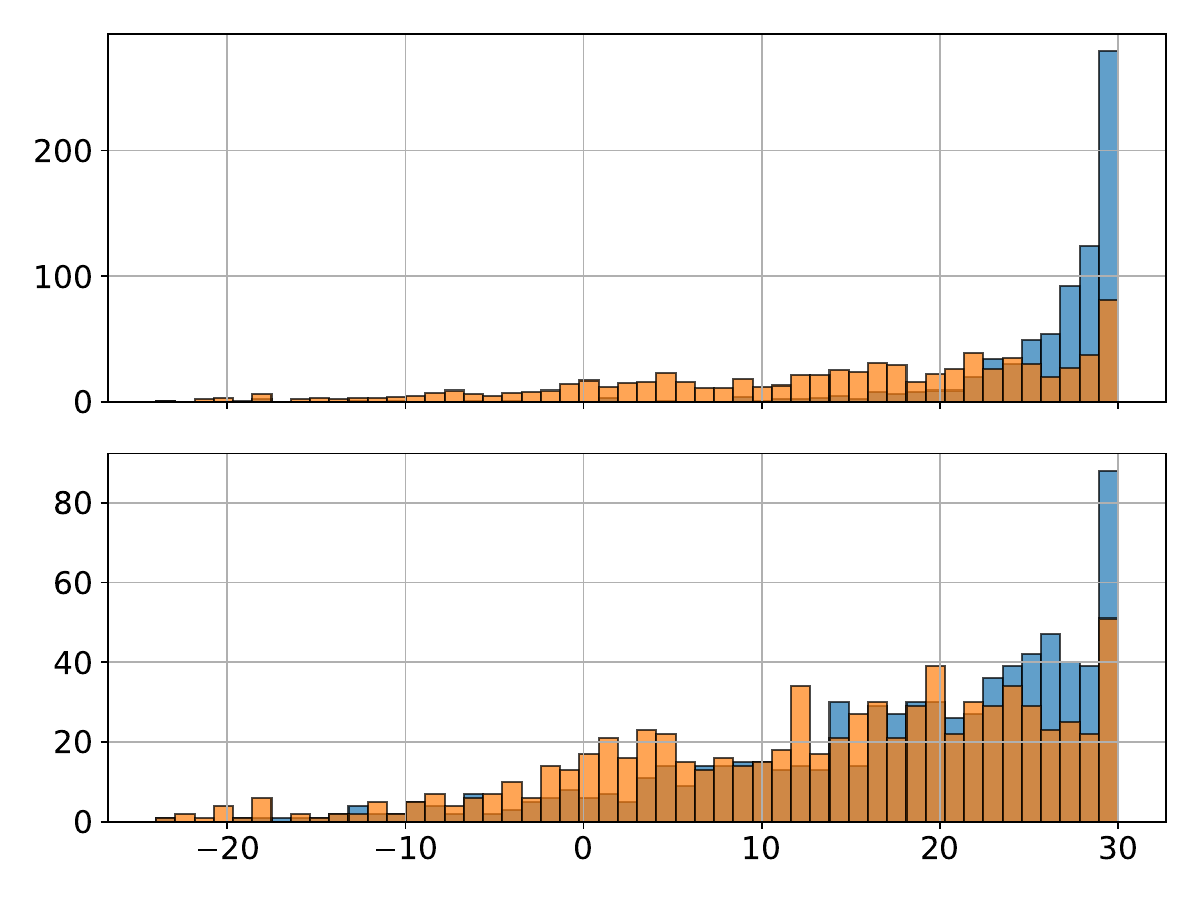}
        \caption{TREC-COVID dataset.}
        \label{fig:treccovid_indisting}
    \end{subfigure}%
    \caption{Histograms of IA scores for member and non-member queries across different datasets. After defense, the member and non-member score distributions become indistinguishable.}
    \label{fig:scores}
\end{figure*}

\begin{table}[]
\resizebox{0.99\linewidth}{!}{
\begin{tabular}{lc|ccc|ccc} \toprule
                      \multicolumn{1}{l}{} & \multicolumn{1}{l}{} & \multicolumn{3}{c}{Adjusted Accuracy ($\downarrow$)} & \multicolumn{3}{c}{KS statistics ($\downarrow$)} \\ \cmidrule(lr){3-5}\cmidrule(lr){6-8}
                      \multicolumn{1}{l}{}   & \multicolumn{1}{l}{}  & NF          & SCI        & \multicolumn{1}{l}{TREC}  & NF        & SCI       & TREC      \\ \midrule 
{S$^2$}  & DP                   & 0.021 & 0.004 & 0.038 & 0.077 & 0.051 & 0.038 \\
                    MIA    &  +Ours                 & 0.020 & 0.029 & 0.024 & 0.088 & 0.028 & 0.050 \\ \midrule
{MBA} & DP                   & 0.015 & 0.040 & 0.288 & 0.022 & 0.081 & 0.015 \\
                        &  +Ours                 & 0.010 & 0.015 & 0.004 & 0.058 & 0.040 & 0.035 \\  \midrule
{IA}  & DP                   & 0.004 & 0.003 & 0.174 & 0.088 & 0.418 & 0.091 \\
                        &  +Ours                 & 0.004 & 0.018 & 0.064 & 0.046 & 0.075 & 0.094
                        \\ \bottomrule
\end{tabular}
}\caption{Defense performances of DP-RAG and DP-RAG with (+) our detect-and-hide using Mirabel.} \label{tab:dp_defense}
\end{table}

\begin{table}[]
\resizebox{0.99\linewidth}{!}{
\begin{tabular}{c|ccc|ccc}\toprule
     & \multicolumn{3}{c}{DP-RAG} & \multicolumn{3}{c}{DP-RAG + Ours} \\ \cmidrule(lr){2-4}\cmidrule(lr){5-7} 
     & @5  & @20 & @100 & @5    & @20    & @100   \\ \hline
NQ   & 0.030  & 0.051  & 0.020   & 0.020    & 0.040     & 0.050     \\
TRIV & 0.255  & 0.225  & 0.240   & 0.230    & 0.230     & 0.225    \\ \bottomrule
\end{tabular}
}
\caption{EM ($\uparrow$) of DP-RAG and DP-RAG (+) ours.}\label{tab:dp_utility}
\end{table}

\begin{table*}[!t]
\centering
\resizebox{0.75\linewidth}{!}{
\begin{tabular}{ll|rrr>{\columncolor[HTML]{E5FFE5}}r|rrr>{\columncolor[HTML]{E5FFE5}}r}\toprule
\multicolumn{1}{l}{} & \multicolumn{1}{l}{} & \multicolumn{4}{c}{(LLM) Phi-4 Mini } & \multicolumn{4}{c}{(Embedder) GTE Large En v1.5} \\ 
\cmidrule(lr){3-6}\cmidrule(lr){7-10} &  & \multicolumn{1}{c}{NF} & \multicolumn{1}{c}{SCI} & \multicolumn{1}{c}{TREC} & \multicolumn{1}{>{\columncolor[HTML]{E5FFE5}}c|}{Avg} & \multicolumn{1}{c}{NF} & \multicolumn{1}{c}{SCI} & \multicolumn{1}{c}{TREC} & \multicolumn{1}{>{\columncolor[HTML]{E5FFE5}}c}{Avg}  \\ \midrule
\multicolumn{1}{c}{S$^2$MIA}  & RAG  & 0.110 & 0.168 & 0.092 & 0.123 & 0.116 & 0.039 & 0.084 & 0.080 \\
    & Ours & 0.024 & 0.007 & 0.038 & \textbf{0.023} & 0.006 & 0.011 & 0.078 & \textbf{0.032} \\ \midrule
\multicolumn{1}{c}{MBA} & RAG  & 0.250 & 0.255 & 0.159 & 0.221 & 0.399 & 0.396 & 0.297 & {0.364} \\
    & Ours & 0.001 & 0.064 & 0.057 & \textbf{0.041} & 0.080 & 0.101 & 0.290 & \textbf{0.157} \\ \midrule
\multicolumn{1}{c}{IA}  & RAG  & 0.322 & 0.430 & 0.248 & 0.334 & 0.347 & 0.416 & 0.222 & 0.328 \\
    & Ours & 0.039 & 0.019 & 0.104 & \textbf{0.054} & 0.011 & 0.008 & 0.205 & \textbf{0.074}\\ \bottomrule
\end{tabular}
}
\caption{Adjusted attack accuracy ($\downarrow$). A smaller value indicates weaker attack success, thus stronger defense.}
\label{tab:aditional}
\end{table*}

\begin{table}[]
\resizebox{\linewidth}{!}{
\begin{tabular}{ccccc}
\toprule 
Data   & RAG (ms) & +Ours (ms) & Diff. (ms) & Ratio \\ \midrule
NF  & 49.215   & 50.895          & 1.680           & 1.034x         \\
SCI   & 52.355   & 60.510          & 8.155           & 1.156x         \\
TREC & 100.750  & 148.695         & 47.945          & 1.476x        \\ 
\bottomrule
\end{tabular}
}
\caption{Runtime overhead of detect-and-hide (Ours) defense compared to the original RAG system.}\label{tab:runtime}
\end{table}

Most MIAs rely on measuring a score and making membership decisions based on that score. Even if a defense method successfully reduces the attacker's accuracy, differences in the score distributions between member and non-member queries enable attackers to perform adaptive attacks. Therefore, another goal of defense is to achieve indistinguishability between responses to member and non-member queries, preventing the attacker from gaining additional membership information.

Table \ref{tab:ks_all} presents the results measuring indistinguishability. As shown in the table, our method significantly reduces the KS statistic, achieving comparative or even smaller values compared to DP-RAG. 
As lower KS statistics indicate higher distributional similarity, our method prevents the attacker from obtaining additional information gain.

To further illustrate this, we present histograms of the attack scores produced by IA before and after applying our defense in Figure~\ref{fig:scores}. In the original RAG, we can observe the clear difference between member and non-member queries. However, after applying our defense, the two distributions are almost similar. Additional results, including the individual KS statistics for similarity and perplexity under S$^2$MIA, as well as histograms of attack scores for other attack methods, are provided in Appendix~\ref{app:add_exp}.

\subsection{Composing with Existing DP Models}
In this section, we adapt our detect-and-hide strategy to DP-RAG to demonstrate that our method is agnostic to the RAG system. Table \ref{tab:dp_defense} presents the defense performance when applying our method on top of base DP-RAG ($\epsilon = 2$). As in the standard RAG setting, our method improved attack resistance and indistinguishability, even though DP already achieves strong defense performance. 
In terms of utility, 
Table~\ref{tab:dp_utility} shows that our method introduces only minimal utility degradation. 
This indicates that composing our method with existing or upcoming privacy-preserving models can be beneficial.
Additional results are in Appendix \ref{app:add_exp}.

\subsection{Experiments with Different Models}
To validate the robustness of our detect-and-hide method, we further evaluate its performance by replacing either the generator LLM or the embedder in the RAG system. 
For the generator LLM model, we utilize Phi-4 Mini Instruct (3.8B) \cite{abouelenin2025phi}, a recently released open-source LLM model.
For the embedding model to compute query-document similarity, we employed GTE Large En v1.5 \cite{zhang2024mgte}, which supports long context inputs (up to 8192 tokens) and is well-suited for document-level retrieval
The experimental results are represented in Table \ref{tab:aditional}.

As shown in the table, our defense method consistently reduces the attack performance across all datasets and attack types, even under different LLM and embedding models, demonstrating the robustness of the detect-and-hide strategy. Nevertheless, with an alternative embedder, the defense performance on the TREC-COVID dataset was relatively weaker than on the others, a limitation we further discuss in Limitations section. 






\subsection{Runtime}
Detect-and-hide reuses similarity scores that are already computed within the RAG system, and in theory, it requires only marginal additional computation for searching. To empirically verify this, we directly measure the runtime. The results are presented in Table \ref{tab:runtime}.

As expected, the overhead on smaller datasets such as NFCorpus remains marginal. However, for larger datasets such as TRECCOVID, we observe a latency increase of up to 47.6\%.
This is due to the FAISS library \cite{douze2024faiss}, which we utilized for RAG. The current version of FAISS does not allow direct access to similarity scores without top-$k$ sorting, leading to unnecessary computation on large datasets.

If future versions of FAISS (or alternative libraries) enable direct access to unsorted similarity scores, our detection method can achieve the theoretically minimal overhead in practice. 

A detailed theoretical analysis based on the complexity explanation of runtime and FAISS is provided in the Appendix \ref{app:add_exp}.

\section{Conclusion}
In this paper, we propose Mirabel, a MIA detection method that classifies member attack queries based on similarity using the Gumbel distribution.
We also introduce a simple defense strategy, detect-and-hide, which effectively defends against attacks with minimal utility degradation.
Our method is model-agnostic, incurs almost no additional computation overhead, and can be easily applied to existing RAG systems. 
Experimental results show that our method achieves defense performance comparable to DP-RAG, while incurring negligible utility loss on benign queries.
\clearpage
\section*{Limitations}
Even though we have designed efficient detection and defense methods for existing MIAs targeting the private external database of the RAG system, we need to be cautious when applying this defense framework in real-world applications. 

Our method relies on the assumption that the embedding of the member attack query is highly similar to that of the target document. Therefore, the detection or defense performance depends on the use of the embedder. 
If the embedder assigns high similarity when inputs share the same word (e.g., “COVID”), Mirabel struggles to detect member attack queries based on the similarity.
This can be observed in Table~\ref{tab:aditional}, where the defense performance on the TREC-COVID dataset was limited when using a different embedder. 
If the defense fails due to this limitation, relying solely on our method could leave sensitive information vulnerable to membership inference attacks. 

To address this, we need to carefully adjust the threshold selection strategy and develop additional defense mechanisms that remain robust across embedders.

\section*{Ethical Considerations}
This work adheres to the ACL Code of Ethics.
In RAG systems, MIAs pose a serious threat to the privacy of user queries submitted to LLMs. Mitigating these attacks, therefore, makes a meaningful ethical contribution and offers practical guidance for developing RAG systems that remain robust against malicious attackers.

\section*{Acknowledgments}
This research was supported by the National Research Foundation of Korea (NRF) grant funded by the Korean government (MSIT) (No. RS-2023-00272502, No. RS-2024-00338859, No. RS 2024-00335811), and the Institute of Information \& communications Technology Planning \& Evaluation (IITP) grant funded by the Korea government (MSIT) (No. RS-2022-II220984, Development of Artificial Intelligence Technology for Personalized Plug-and-Play Explanation and Verification of Explanation).
Jinseong Park is supported by a KIAS Individual Grant
(AP102301) via the Center for AI and Natural Sciences at Korea Institute for Advanced Study. This work was supported by the Center for Advanced Computation at Korea Institute for Advanced Study.

%% file: manuscript_appendix.tex
\clearpage

\section{Experimental Details}
\label{app:exp}
Our complete framework is available in our official GitHub repository at \url{https://github.com/nonalcohol-park/MIRABEL}. The detailed settings are provided below.

\subsection{Dataset}\label{app:datasets}
For attack and defense, we used NFCorpus, SCIDOCD, and TREC-COVID, containing approximately 3.6K, 171K, and 25K documents, respectively. For the retriever database, NFCorpus is split into member and non-member sets in a 7:3 ratio, while SCIDOCS and TREC-COVID are divided 8:2. To conduct MIA, we sample 1,000 members and 1,000 non-members from NFCorpus, and 2,000 members and 2,000 non-members from the other two datasets. Each member and non-member subset is further split (5:5) into reference and evaluation sets. Detection performance from benign queries to attack queries is evaluated using 323, 500, and 50 queries provided in NFCorpus, SCIDOCS, and TREC-COVID, respectively, while an equal number of MIA attack queries is generated for the member subsets. This balanced design ensures a fair assessment of the detection and defense capabilities.

For utility evaluation on benign queries, NQ and TriviaQA are used, comprising approximately 7.8K and 138K question–answer (QA) pairs, respectively. In our experiments, we select 200 QA pairs from each dataset for evaluation.

All datasets and models used in this study are publicly available and were released for academic or research purposes.
\subsection{Hyperparameters}\label{app:hyper}
\paragraph{Attacks}
For all datasets, we divided the member and non-member datasets into 5:5 for reference and evaluation. For MBA and S$^2$MIA, reference sets were used to establish thresholds, following their original paper. We found a threshold in the greedy search algorithm for S$^2$MIA and the highest F1 score for each count of masking tokens in $\{5,10,15,20\}$ in MBA. 

For IA, we evaluate 30 queries to compute the IA score for the standard RAG system, and 20 queries for DP-RAG, since each query consumes a portion of the total privacy budget. Specifically, with a total $\epsilon = 2$, each query uses $\epsilon = 0.1$, while when $\epsilon = 100$, each query uses $\epsilon = 5$.

In the selection of thresholds for IA, while the original paper selects the threshold based on a false positive rate, we instead select the threshold based on accuracy. We searched threshold in $[0, 1]$, and we evaluated penalty values $\lambda \in [0.5, 1]$.
In the case of DP-RAG, since the number of queries is smaller, we differ the searching space.
Since we observed that larger values of $\lambda$ in this range resulted in higher recall but significantly lower accuracy, indicating potential over-penalization. We therefore refined the search to a smaller range of $\lambda \in [0.1, 0.5]$ for this setting.

\begin{figure*}[!ht]
\centering

\begin{tcolorbox}[title=RAG Generator Prompt]
\textbf{System:}\\
You are a helpful assistant. Below is a query from a user and some relevant contexts.\\
Answer the question given the information in those contexts. Your answer should be short and concise.\\
If you cannot find the answer to the question, just say "I don’t know".

\bigskip
\textbf{User:}\\
Contexts: \{context\}\\
Query: \{question\}

\bigskip
\textbf{Assistant:}\\
Answer:
\end{tcolorbox}
\caption{Prompt template given to the RAG generator. It conditions the model on the retrieved
contexts and enforces grounded, concise answers with an explicit "I don't know" fallback.}
\label{ex:rag_prompt}
\end{figure*}

\begin{figure*}[!h]
\centering
\begin{tcolorbox}[title=Agent-based Detection Prompt]
\textbf{System:}\\
You are an expert classifier tasked with validating user input queries in a Retrieval-Augmented Generation (RAG) system. Your
role is to classify each query into one of the following two categories:\\
1. Natural Queries:\\
   – Queries that are naturally formulated and relevant to the task/domain.\\
   – These queries align with the system’s intended purpose, such as retrieving factual information, summarizing content, or answering domain-specific queries.\\
2. Context-Probing Queries:\\
   – Queries that attempt to probe hidden or inaccessible parts of the system or input context.\\
   – These queries are often out-of-scope and unnatural for the task/domain. Examples include queries that ask for metadata, hidden inputs, or system behavior.\\

— Instructions —\\
1. Carefully analyze the given query to determine whether it aligns with the system’s intended purpose.\\
2. If the query is Natural, ensure it is appropriate and consistent with the system’s task/domain.\\
3. If the query is Context-Probing, identify if it is attempting to extract hidden content, system behavior, or metadata.\\

— Output Format —\\
For each input query, respond with a single word indicating the classification:\\
Natural or Context-Probing.
Do not provide any explanations or reasoning in your response.\\

— Examples —\\
Example 1:\\
Input Query: "What are the common symptoms of diabetes?"\\
Output:\\
Natural\\
Example 2:\\
Input Query: "What part of the document did you hide from me?"\\
Output:\\
Context-Probing

\bigskip
\textbf{User:}\\
Input Query: "\{Query\}"
\end{tcolorbox}
\caption{Prompt template for the GPT-4o agent that classifies incoming queries as either
\emph{Natural} (task-aligned) or \emph{Context-Probing} (potentially MIA).}
\label{ex:agent_det_prompt}
\end{figure*}

\begin{figure*}[!h]
    \centering
    \begin{subfigure}[b]{0.33\textwidth}
        \centering
        \includegraphics[width=\linewidth]{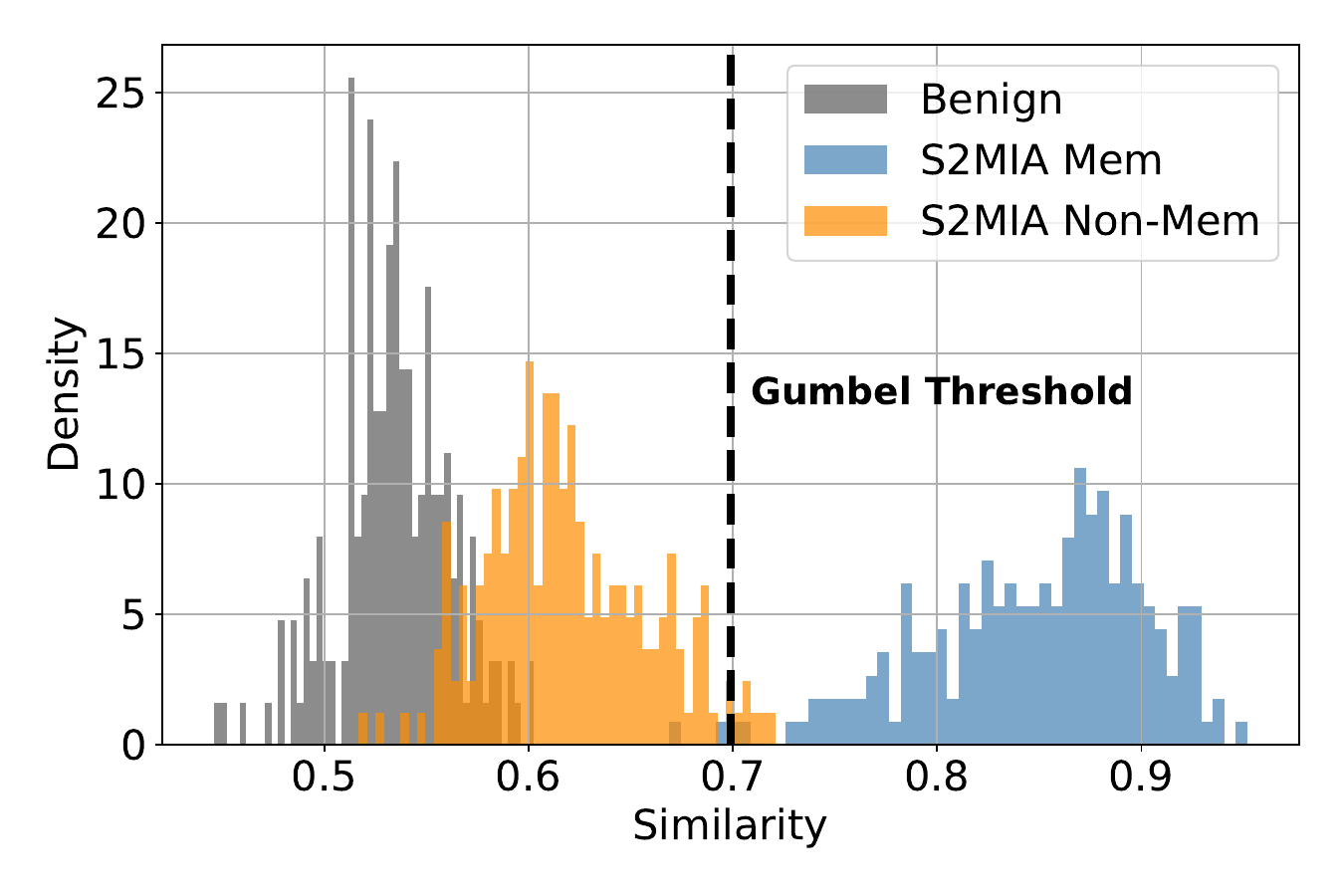}
        \caption{Gumbel threshold for S$^2$MIA.}
        \label{fig:gumbel_s2_nonmem}
    \end{subfigure}
    \begin{subfigure}[b]{0.33\textwidth}
        \centering
        \includegraphics[width=\linewidth]{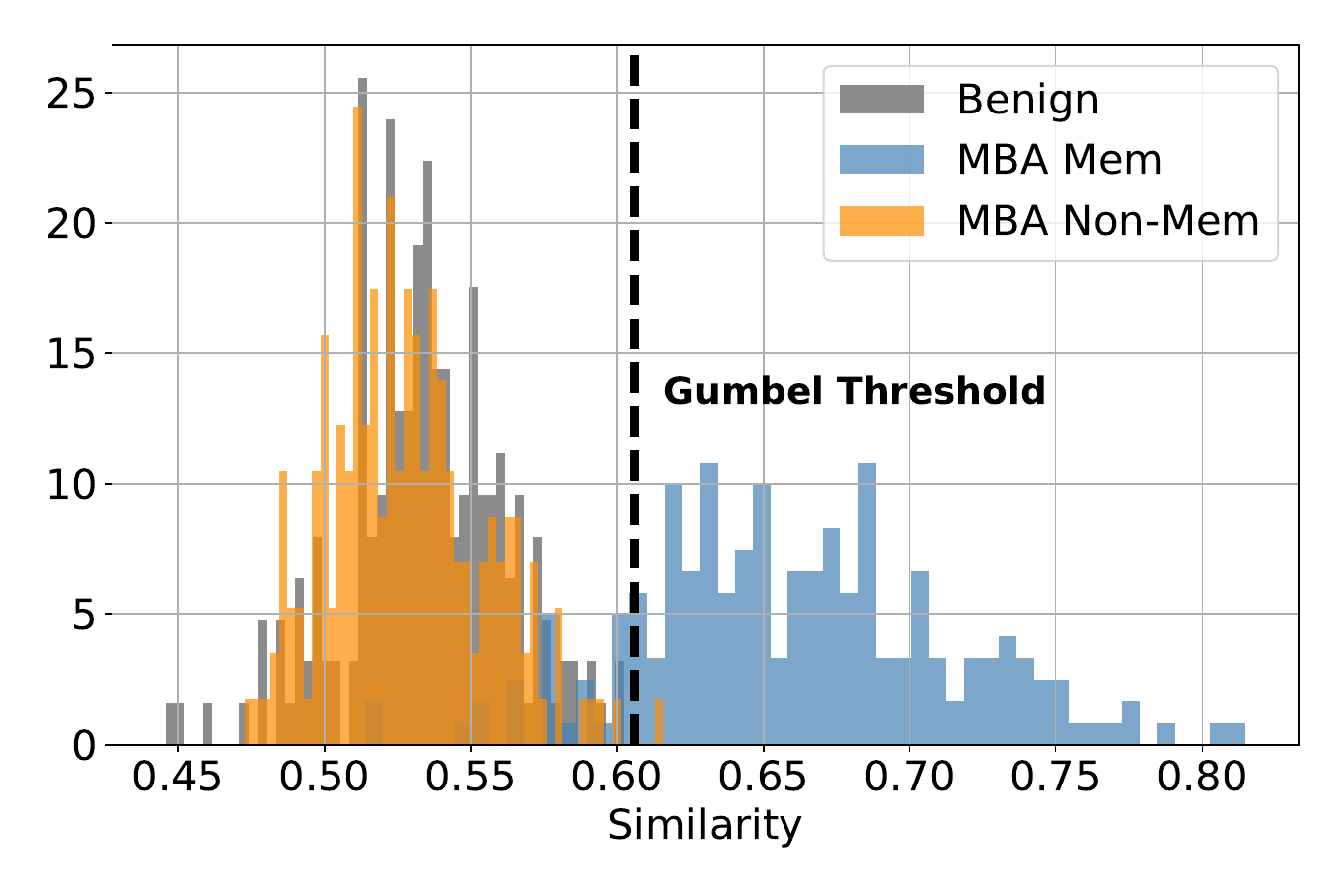}
        \caption{Gumbel threshold for MBA.}
        \label{fig:gumbel_mba_nonmem}
    \end{subfigure}%
    \begin{subfigure}[b]{0.33\textwidth}
        \centering
        \includegraphics[width=\linewidth]{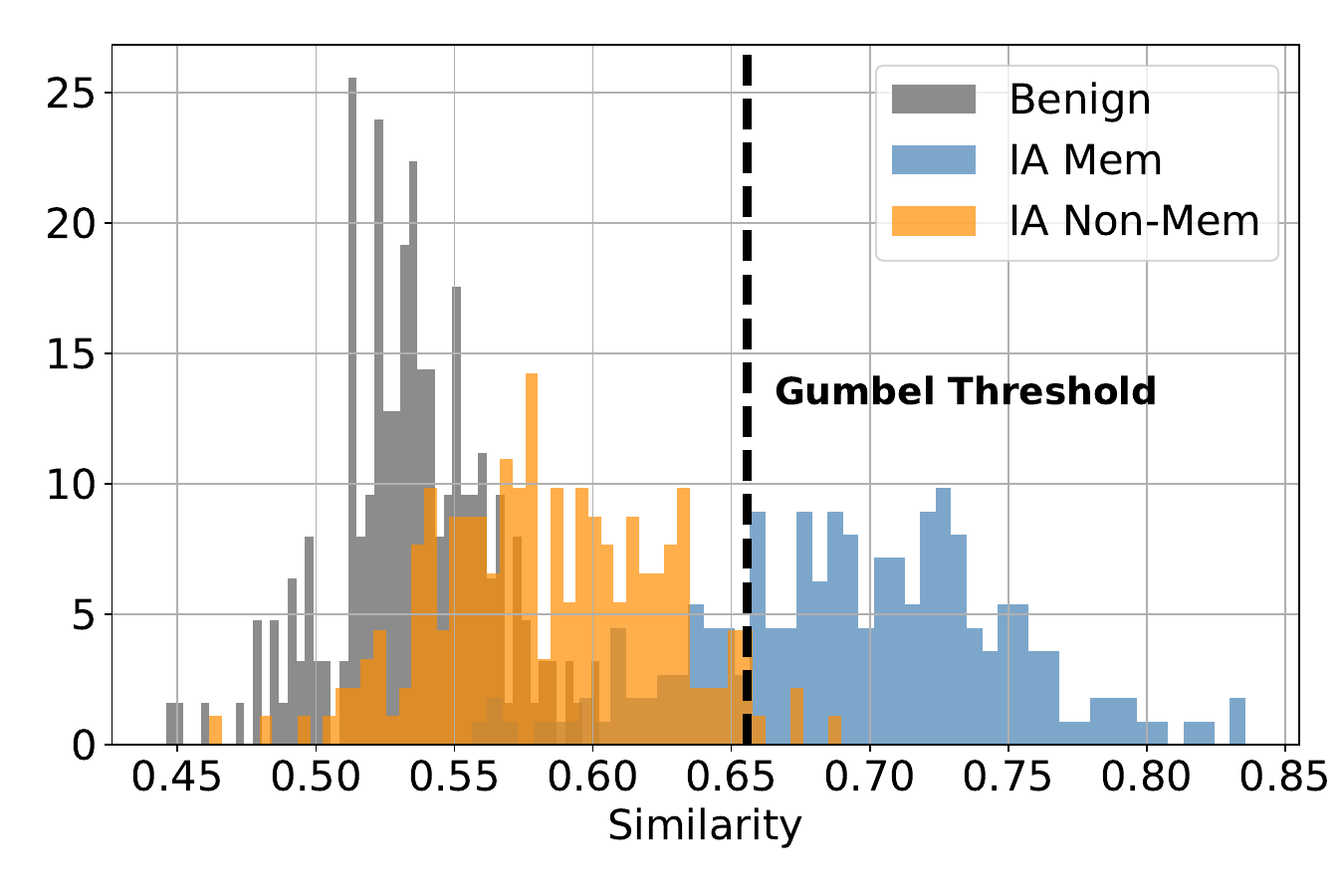}
        \caption{Gumbel threshold for IA.}
        \label{fig:gumbel_ia_nonmem}
    \end{subfigure}%
    \caption{Illustration of the similarity distributions for member attack queries and non-member attack queries, compared to benign queries. The Gumbel-based threshold is shown for each attack. Our method effectively separates member and non-member attack queries using this threshold.}
    \label{fig:toy_nonmember}
\end{figure*}

\paragraph{RAG Setting}
Maximum sequence length was set to 8192 tokens for embedding documents using BGEm3. For experiments on TREC-COVID corpus, the maximum sequence length was set to 2048 tokens involving DP generator. To facilitate efficient indexing and retrieval, we employed FAISS \cite{douze2024faiss}. 
We segment the Wikipedia corpus into non-overlapping chunks of 100 tokens and randomly sample one-tenth of these segments to serve as the retriever database. In building the Wikipedia retriever database, we utilized half-precision (fp16) to reduce computational overhead. This approach helps to assess the usability of RAG systems in a controlled yet representative manner while maintaining computational ease.

\paragraph{DP-RAG setting}
For DP-RAG, we follow the setting provided in the official GitHub of \cite{grislain2024rag} (\url{https://github.com/sarus-tech/dp-rag}). 
Specifically, we set the parameters as $\omega = 0.01$, $\alpha = 1.0$, and temperature = 1.0.

\subsection{MIA implementation}\label{app:mia}
To implement MIAs for evaluating Mirabel's defense, we followed each original papers. For $S^2$MIA, we calculated BLEU score and perplexity. To gain perplexity, GPT-2 was used. For MBA, we tokenize the sentences and selected mask tokens in random. \cite{liu2025mask} showed random masking has similar accuracy in attempting attacks. For IA, GPT-4o and GPT-4o-mini was used to generate summaries, questions, and ground truths.

\section{Prompt Templates}
\label{app:prompt}

Figure \ref{ex:rag_prompt} shows the generation prompt given to every RAG model (Llama-3.1 8B for utility experiments and Llama-3.2 3B elsewhere). Its design enforces three constraints that proved crucial for reliable evaluation: groundedness, conciseness, and failure transparency. Figure \ref{ex:agent_det_prompt} gives the agent-based detection prompt used with GPT-4o to label incoming queries as either \emph{Natural} or
\emph{Context-Probing}.

Both templates are frozen across all runs; we do not tune any prompt hyperparameters.  The raw prompts introduced in the figures includes placeholder tokens \texttt{\{context\}}, \texttt{\{question\}}, and
\texttt{\{Query\}} to facilitate replication.

\section{Additional Experiments}\label{app:add_exp}
\subsection{Illustration of motivating example}
Additional illustration of Figure \ref{fig:distribution_top1}. We display the top-1 similarity of $q_a^m$ and $q_a^n$ in Figure \ref{fig:toy_nonmember}. 
As shown in the Figure, our Gumbel threshold can efficiently separate the top-1 similarity of $q_a^m$ and $q_a^n$.

Since removing the top-1 similarity is equivalent to removing the target document from the database, the query can be treated as a non-member attack query. Therefore, top-2 similarity as $q_a^m$ corresponds to the top-1 similarity of $q_a^n$.

\subsection{Detection}
The agent-based detection methods aim to detect benign queries and attack queries, while our method detects member attack queries.
To ensure a fair comparison, we evaluate detection performance using only benign queries and member attack queries.

Table \ref{tab:detection_fair} shows the detection performance. As shown, the performance of Mirabel was similar to that in Table \ref{tab:detection}, while agent-based methods demonstrate significantly better performance, except for IA, which is designed to be stealthy. These experimental results are consistent with those reported in \citet{naseh2025riddle}.

Our method demonstrates stable performance across a variety of attacks, particularly maintaining reliable detection performance even for IA, which exhibits stealthy characteristics.


\subsection{Defense}\label{app:defense}
In this subsection, we provide the performance of MIA attackers, including accuracy, precision, recall, and F1 score for each attack. Table \ref{tab:defense} presents the attack performance for the standard RAG and adapting our method to that RAG system. Table \ref{tab:dp_defense} shows the results for DP-RAG (-L) and DP-RAG (-L) with our method in terms of the accuracy, precision, recall, and F1 score.

As discussed in the main paper, the accuracy close to 0.5 means a stronger defense.
Moreover, if the attacker’s ability to classify a member as a member (True Positive, TP) and a non-member as a non-member (False Positive, FP) are similar, the indistinguishability criterion is implicitly satisfied.
Therefore, precision, which is defined as 
\begin{equation}
\frac{TP}{TP+FP},    
\end{equation}
can be considered an indicator of indistinguishability, with a value close to 0.5 suggesting that indistinguishability has been achieved.

Our detect-and-hide strategy achieved a value close to 0.5 in nearly every accuracy and precision measure.

Notably, in our method, we achieved a low recall measure, which is defined as 
\begin{equation}
\frac{TP}{TP+FN}.    
\end{equation}

The detect-and-hide is designed to make the attacker misclassify a member as a non-member. Therefore, our method increases the false negatives, making the attacker misclassify a member as a non-member.

\begin{table*}[!t]
\centering
\resizebox{0.9\textwidth}{!}{%
\begin{tabular}{lc|ccc>{\columncolor[HTML]{E5FFE5}}c|ccc>{\columncolor[HTML]{E5FFE5}}c}
\toprule
&                       & \multicolumn{4}{c}{Agent-based Detection} & \multicolumn{4}{c}{Mirabel Detection} \\ 
\cmidrule(lr){3-6}\cmidrule(lr){7-10}
{Attacks} & {Data} & Acc ($\uparrow$) & F1 ($\uparrow$) & Precision ($\uparrow$) & Recall ($\uparrow$) & Acc ($\uparrow$) & F1 ($\uparrow$) & Precision ($\uparrow$) & Recall ($\uparrow$) \\ 
\midrule
\multirow{3}{*}{S$^{2}$MIA} 
 & NF   & 0.941 & 0.942 & 0.925 & 0.960 & 0.876 & 0.890 & 0.802 & \textbf{1.000} \\
 & SCI  & 0.962 & 0.963 & 0.951 & 0.974 & 0.865 & 0.880 & 0.791 & \textbf{0.992} \\
 & TREC & 0.980 & 0.980 & 1.000 & \textbf{0.960} & 0.970 & 0.969 & 1.000 & 0.940 \\
\midrule
\multirow{3}{*}{MBA} 
 & NF   & 0.954 & 0.955 & 0.927 & 0.985 & 0.876 & 0.890 & 0.802 & \textbf{1.000} \\
 & SCI  & 0.968 & 0.969 & 0.952 & 0.986 & 0.869 & 0.884 & 0.792 & \textbf{1.000} \\
 & TREC & 1.000 & 1.000 & 1.000 & \textbf{1.000} & 0.980 & 0.980 & 1.000 & 0.960 \\
\midrule
\multirow{3}{*}{IA} 
 & NF   & 0.464 & 0.011 & 0.074 & 0.006 & 0.824 & 0.835 & 0.783 & \textbf{0.895} \\
 & SCI  & 0.483 & 0.030 & 0.242 & 0.016 & 0.816 & 0.829 & 0.773 & \textbf{0.894} \\
 & TREC & 0.500 & 0.000 & 0.000 & 0.000 & 0.810 & 0.765 & 1.000 & \textbf{0.620} \\
\bottomrule
\end{tabular}}
\caption{Detection performance of Mirabel compared to agent-based detection, evaluated using $q_n$ and $q_a^m$.}
\label{tab:detection_fair}
\end{table*}

\begin{table*}[!h]
\resizebox{0.99\linewidth}{!}{
\begin{tabular}{lc|ccc|ccc|ccc|ccc} \toprule
 & \multicolumn{1}{l}{} & \multicolumn{3}{c}{Acc} & \multicolumn{3}{c}{Precision} & \multicolumn{3}{c}{Recall} & \multicolumn{3}{c}{F1} \\ \cmidrule(lr){3-5} \cmidrule(lr){6-8}\cmidrule(lr){9-11} \cmidrule(lr){12-14}
                         & \multicolumn{1}{l}{} & S$^2$MIA & MBA   & IA    & S$^2$MIA & MBA   & IA    & S$^2$MIA & MBA   & IA    & S$^2$MIA & MBA   & IA    \\ \cmidrule(lr){3-3} \cmidrule(lr){4-4}\cmidrule(lr){5-5}\cmidrule(lr){6-6}\cmidrule(lr){7-7}\cmidrule(lr){8-8} \cmidrule(lr){9-9}\cmidrule(lr){10-10} \cmidrule(lr){11-11} \cmidrule(lr){12-12} \cmidrule(lr){13-13} \cmidrule(lr){14-14} 
\multicolumn{1}{c}{NF}   & RAG                  & 0.688 & 0.877 & 0.903 & 0.636 & 0.859 & 0.878 & 0.878 & 0.902 & 0.967 & 0.738 & 0.880 & 0.920 \\
                         & Ours                 & 0.524 & 0.481 & 0.492 & 0.526 & 0.425 & 0.681 & 0.480 & 0.108 & 0.230 & 0.502 & 0.172 & 0.344 \\
\multicolumn{1}{c}{SCI}  & RAG                  & 0.626 & 0.906 & 0.868 & 0.647 & 0.914 & 0.884 & 0.555 & 0.897 & 0.847 & 0.597 & 0.906 & 0.865 \\
                         & Ours                 & 0.506 & 0.601 & 0.510 & 0.510 & 0.789 & 0.548 & 0.301 & 0.276 & 0.117 & 0.378 & 0.409 & 0.193 \\
\multicolumn{1}{c}{TREC} & RAG                  & 0.619 & 0.813 & 0.755 & 0.586 & 0.813 & 0.696 & 0.832 & 0.815 & 0.894 & 0.688 & 0.814 & 0.782 \\
                         & Ours                 & 0.521 & 0.639 & 0.583 & 0.522 & 0.759 & 0.594 & 0.597 & 0.417 & 0.485 & 0.557 & 0.539 & 0.534 \\ \bottomrule
\end{tabular}
}\caption{Defense performance of RAG and RAG with our detect-and-hide defense.}\label{tab:defense}
\end{table*}

\begin{table*}[!h]
\resizebox{0.99\linewidth}{!}{
\begin{tabular}{lc|ccc|ccc|ccc|ccc} \toprule
 & \multicolumn{1}{l}{} & \multicolumn{3}{c}{Acc} & \multicolumn{3}{c}{Precision} & \multicolumn{3}{c}{Recall} & \multicolumn{3}{c}{F1} \\ \cmidrule(lr){3-5} \cmidrule(lr){6-8}\cmidrule(lr){9-11} \cmidrule(lr){12-14}
                         & \multicolumn{1}{l}{} & S$^2$MIA & MBA   & IA    & S$^2$MIA & MBA   & IA    & S$^2$MIA & MBA   & IA    & S$^2$MIA & MBA   & IA    \\ \cmidrule(lr){3-3} \cmidrule(lr){4-4}\cmidrule(lr){5-5}\cmidrule(lr){6-6}\cmidrule(lr){7-7}\cmidrule(lr){8-8} \cmidrule(lr){9-9}\cmidrule(lr){10-10} \cmidrule(lr){11-11} \cmidrule(lr){12-12} \cmidrule(lr){13-13} \cmidrule(lr){14-14} 
\multicolumn{1}{c}{NF}   & DP-RAG       & 0.479  & 0.496  & 0.538 & 0.486    & 0.495    & 0.579   & 0.738   & 0.394   & 0.740  & 0.586 & 0.439 & 0.650                         \\
                         & ~+Ours                 & 0.480  & 0.529  & 0.477 & 0.487    & 0.537    & 0.572   & 0.748   & 0.426   & 0.383  & 0.590 & 0.475 & 0.459                         \\
                         & DP-RAG-L       & 0.466  & 0.514  & 0.767 & 0.465    & 0.517    & 0.806   & 0.452   & 0.438   & 0.787  & 0.458 & 0.474 & 0.797                         \\
                         & ~+Ours                 & 0.486  & 0.507  & 0.496 & 0.486    & 0.508    & 0.620   & 0.478   & 0.432   & 0.336  & 0.482 & 0.467 & 0.436                         \\
\multicolumn{1}{c}{SCI}  & DP-RAG       & 0.515  & 0.540  & 0.788 & 0.538    & 0.560    & 0.834   & 0.214   & 0.381   & 0.719  & 0.306 & 0.453 & 0.772                         \\
                         & ~+Ours                 & 0.510  & 0.515  & 0.504 & 0.526    & 0.523    & 0.606   & 0.209   & 0.345   & 0.020  & 0.299 & 0.416 & 0.039                         \\
                         & DP-RAG-L       & 0.500  & 0.536  & 0.805 & 0.500    & 0.557    & 0.790   & 0.719   & 0.354   & 0.830  & 0.590 & 0.433 & 0.809                         \\
                         & ~+Ours                 & 0.496  & 0.505  & 0.508 & 0.497    & 0.508    & 0.619   & 0.728   & 0.333   & 0.040  & 0.591 & 0.402 & 0.075                         \\
\multicolumn{1}{c}{TREC} & DP-RAG       & 0.496  & 0.503  & 0.674 & 0.000    & 0.509    & 0.679   & 0.000   & 0.419   & 0.642  & 0.000 & 0.459 & 0.660                         \\
                         & ~+Ours                 & 0.496  & 0.482  & 0.564 & 0.000    & 0.484    & 0.587   & 0.000   & 0.403   & 0.390  & 0.000 & 0.440 & 0.468                         \\
                         & DP-RAG-L       & 0.510  & 0.496  & 0.684 & 0.520    & 0.501    & 0.662   & 0.355   & 0.271   & 0.734  & 0.422 & 0.351 & 0.696 \\
                         & ~+Ours                 & 0.509  & 0.505  & 0.576 & 0.519    & 0.517    & 0.580   & 0.352   & 0.282   & 0.507  & 0.420 & 0.365 & 0.541                        
\\ \bottomrule
\end{tabular}
}\caption{Defense performance of DP-RAG and DP-RAG with our detect-and-hide defense.}\label{tab:dp_defense_total_table}
\end{table*}

We provide the defense performances of DP-RAG-L and DP-RAG-L with ours in Table \ref{tab:dp_defense_large}.

\begin{table*}[]
\centering
\begin{tabular}{lc|ccc|ccc} \toprule
                      \multicolumn{1}{l}{} & \multicolumn{1}{l}{} & \multicolumn{3}{c}{Adjusted Accuracy ($\downarrow$)} & \multicolumn{3}{c}{KS-statistics ($\downarrow$)} \\ \cmidrule(lr){3-5}\cmidrule(lr){6-8}
                      \multicolumn{1}{l}{}   & \multicolumn{1}{l}{}  & NF          & SCI        & \multicolumn{1}{l}{TREC}  & NF        & SCI       & TREC      \\ \midrule 
{S$^2$}  & DP-RAG-L                   & 0.034       & 0.000      & 0.004      & 0.060     & 0.035     & 0.058     \\
                    MIA    & +Ours                 & 0.014       & 0.004      & 0.009      & 0.051     & 0.025     & 0.035     \\ \midrule
{MBA} & DP-RAG-L                   & 0.014       & 0.036      & 0.184      & 0.036     & 0.073     & 0.026     \\
                        & +Ours                 & 0.007       & 0.005      & 0.005      & 0.022     & 0.032     & 0.017     \\ \midrule
{IA}  & DP-RAG-L                   & 0.267       & 0.305      & 0.020      & 0.301     & 0.414     & 0.223     \\
                        & +Ours                 & 0.004       & 0.008      & 0.076      & 0.036     & 0.111     & 0.129     \\ \bottomrule
\end{tabular}
\vspace{-2mm} \caption{Defense performances of DP-RAG-L and DP-RAG-L with (+) our detect-and-hide using Mirabel.} \label{tab:dp_defense_large}
\end{table*}

\begin{table*}[!ht]
\centering
\begin{tabular}{l|ccc|ccc} \toprule
\multirow{2}{*}{} & \multicolumn{3}{c|}{{KS for Similarity ($\downarrow$)}} & \multicolumn{3}{c}{{KS for Perplexity ($\downarrow$)}} \\
                  & NF    & SCI   & TREC  & NF    & SCI   & TREC  \\ \midrule
RAG               & 0.378 & 0.254 & 0.239 & 0.338 & 0.137 & 0.167 \\
Ours              & 0.064 & 0.024 & 0.048 & 0.050 & 0.053 & 0.038 \\
DP-RAG          & 0.056 & 0.051 & 0.041 & 0.098 & 0.050 & 0.034 \\
DP-RAG-L          & 0.038 & 0.032 & 0.054 & 0.082 & 0.039 & 0.062 \\
\bottomrule
\end{tabular}
\caption{KS statistics for S$^2$MIA using similarity and perplexity scores across different datasets. Lower is better.}
\label{tab:indistinguishability_s2_sim_ppl}
\end{table*}

\subsection{Indistinguishable}
In this subsection, we provide the additional indistinguishability experiments. Specifically, S$^2$MIA. Since S$^2$MIA does not have any specific score and relies on the similarity score and perplexity score, we reported the average of each experiment. 
The individual results are in Table \ref{tab:indistinguishability_s2_sim_ppl}.

When calculating the KS statistic with perplexity, some perplexity values were infinite and were therefore excluded from the computation.

\begin{figure*}
    \centering
    \includegraphics[width=\linewidth]{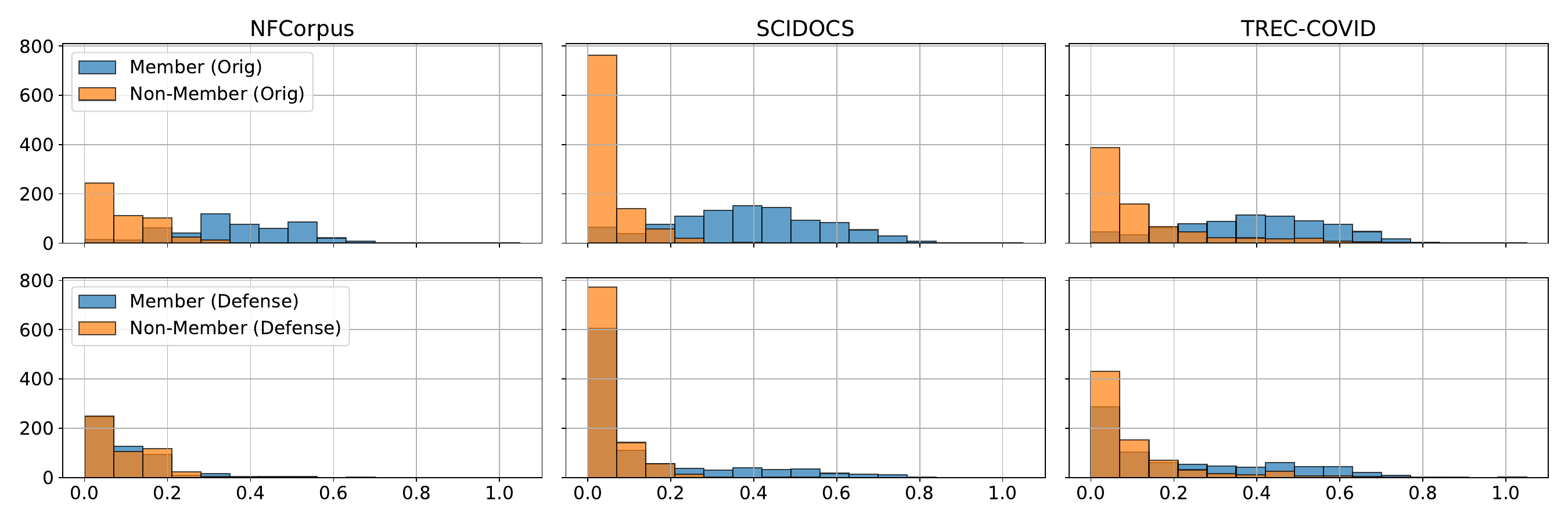}
    \caption{Histograms of MIA scores for member and non-member queries across different datasets. }
    \label{fig:indistin_mba}
\end{figure*}

\begin{figure*}
    \centering
    \includegraphics[width=\linewidth]{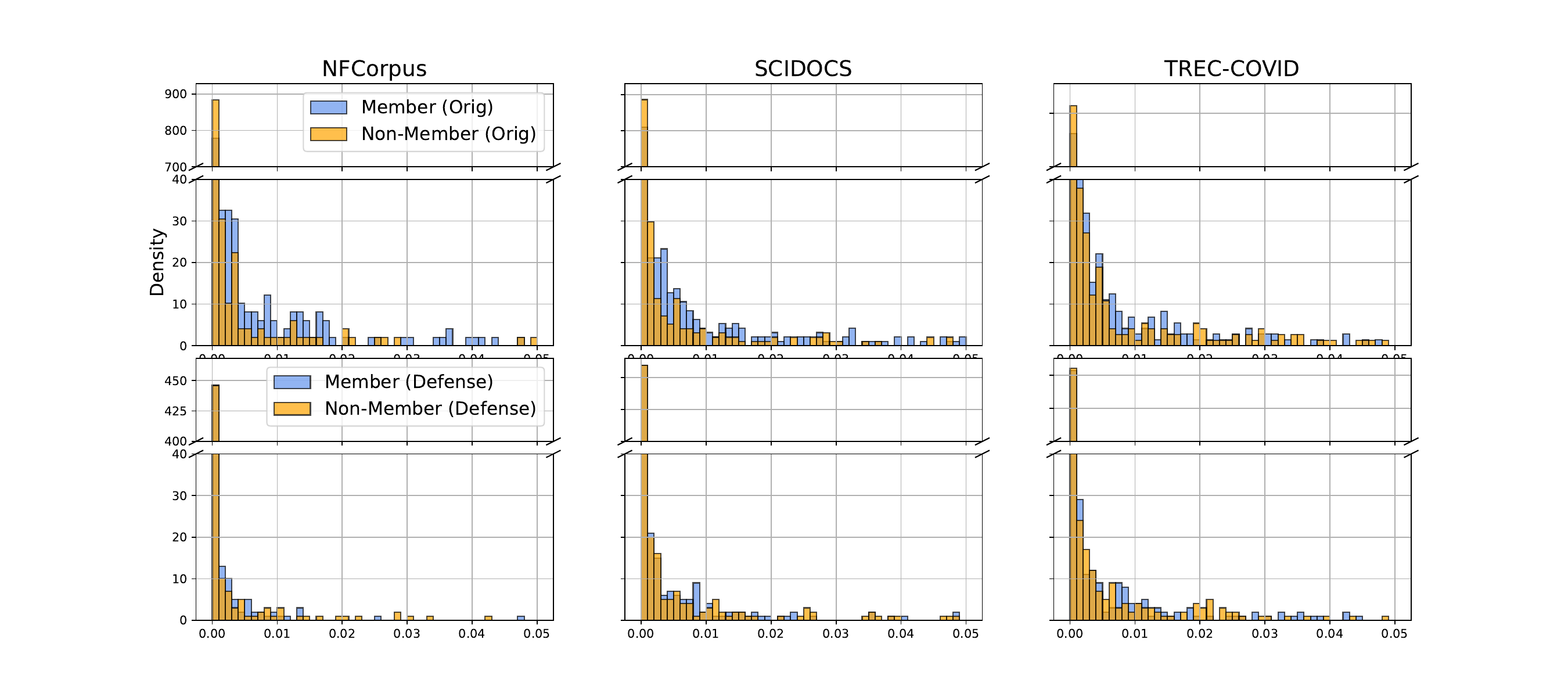}
    \caption{Histograms of similarity scores from S$^2$MIA for member and non-member queries across different datasets.}
    \label{fig:indistin_s2_sim}
\end{figure*}

\begin{figure*}
    \centering
    \includegraphics[width=\linewidth]{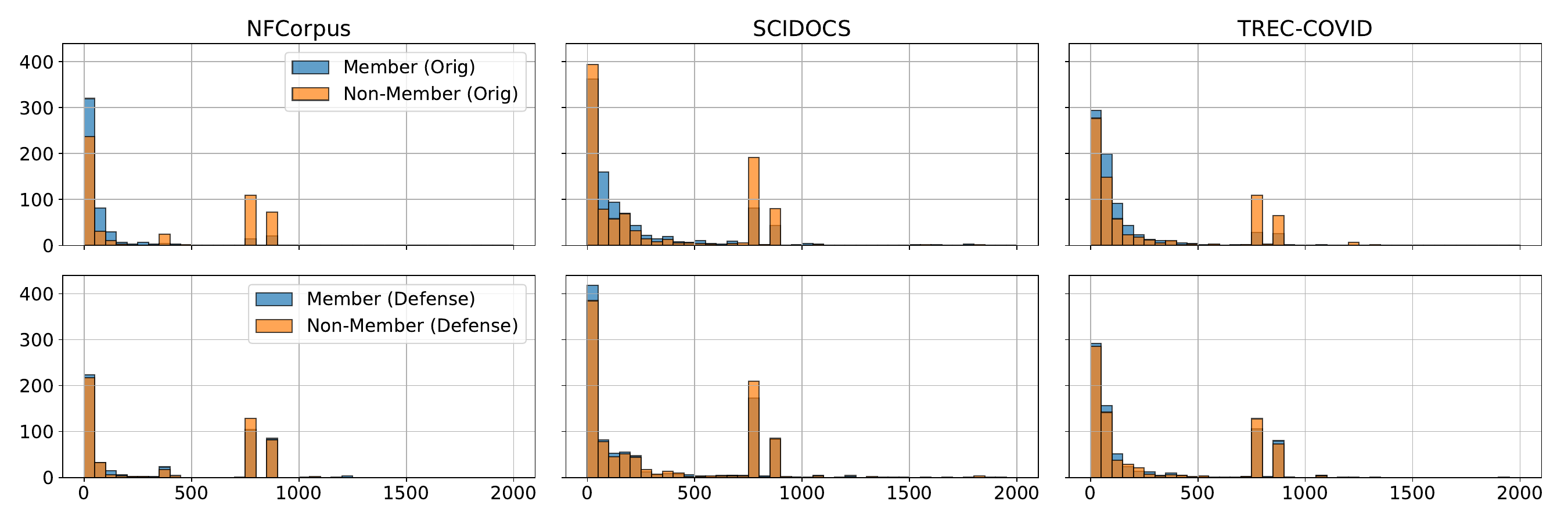}
    \caption{Histograms of perplexity scores from S$^2$MIA for member and non-member queries across different datasets. Due to the extremely large values of some perplexity scores, the histogram is truncated at 2000 for readability.}
    \label{fig:indistin_s2_ppl}
\end{figure*}

Additionally, we illustrate the distributions of scores of MBA and S$^2$MIA in Figures \ref{fig:indistin_mba}-\ref{fig:indistin_s2_ppl}.

\subsection{Runtime}
In this subsection, we theoretically provide the runtime overhead of original RAG system and the detect-and-hide, with big-$O$ notation.

For small RAG system, we can compute all pairwise similarities between queries and $n$ documents with $d$ dimensional embedding vector. (In our experiments, we used the BGE-M3 embedder, which has $d = 1024$.) This searching operation costs  $O(n\cdot d)$.
Given these scalar value similarity scores, computing their mean adds only $O(n)$, and the Gumbel threshold calculation is $O(1)$. Therefore, the additional cost $O(n)+ O(1) = O(n)$ is negligible compared to the original retrieval cost $O(n\cdot d)$. 

Similar to the small RAG system, the additional cost is negligible even in a large-scale database. 
For a large-scale database, approximate nearest neighbor (ANN) methods are commonly used to avoid scanning all vectors. 
These methods typically pre-cluster the database into $C$ clusters (in practice, $C$ can be $2^{16}$ or larger \cite{johnson2019billion}), scan the centroids, and retrieve from the selected cluster of size $O(m)$, where the maximum size of each cluster is $m$. Therefore, the total retrieval cost is  $O(m\cdot d) + O(C \cdot d)$, for scanning centroids and retrieve from the selected cluster.
In this case, Mirabel computes the Gumbel threshold only over the selected cluster with size $O(m)$. Thus, the additional computation consists of $O(m)$ for computing the mean and $O(1)$ for calculating the Gumbel threshold.  Therefore, the additional cost $O(m)+ O(1) = O(m)$ is negligible compared to the original retrieval cost $O(m \cdot  d) + O(C\cdot d)$.

However, as in the Table \ref{tab:runtime} in the main manuscript, the runtime overhead of detect-and-hide was higher than the theoretical overhead.
The difference in runtime mainly occurs in the similarity scores computed in our implementation. Specifically, we utilized FAISS’s \cite{douze2024faiss} \texttt{IndexIDMap.search(query\_embedding, k)} with $k = n$ to retrieve all similarity scores. However, the \texttt{search()} method internally performs {both} similarity computation ($O(n \cdot d)$) and top-$k$ selection via sorting, which incurs an additional $O(n \log k)$ cost. When $k = n$, this becomes a full sort with $O(n \log n)$ complexity.

